\documentclass[journal]{IEEEtran}

\usepackage[utf8]{inputenc}
\usepackage{amsmath}
\usepackage{amssymb}
\usepackage{balance}
\usepackage{graphicx}
\usepackage{booktabs}
\usepackage{color}
\usepackage{cite}
\usepackage{tabu}
\usepackage{algorithm}
\usepackage{algorithmic}
\usepackage[caption = false,font = footnotesize]{subfig}
\usepackage{float}
\usepackage[normalem]{ulem}

\newcommand{\qedsymbol}{\hspace{\fill}\rule{1.5ex}{1.5ex}}

\definecolor{bs}{rgb}{0.0, 0.0, 0.0}

\begin{document}

\title{A Meta-Learning Approach for Training Explainable Graph Neural Networks}
\author{Indro Spinelli, Simone Scardapane and Aurelio Uncini
\thanks{This work was partially supported by the CHIST-ERA grant CHIST-ERA-19-XAI-009. Corresponding author email: indro.spinelli@uniroma1.it. The authors are with the Department of Information Engineering, Electronics and Telecommunications (DIET), Sapienza University of Rome, Italy. Email: \{indro.spinelli, simone.scardapane, aurelio.uncini\}.}\vspace{-.5cm}}

\markboth{Published in IEEE TNNLS Special Issue on Deep Neural Networks for Graphs: Theory, Models, Algorithms and Applications}%
{Spinelli \MakeLowercase{\textit{et al.}}: MATE}

\maketitle

\begin{abstract}
In this paper, we investigate the degree of explainability of graph neural networks (GNNs). Existing explainers work by finding global/local subgraphs to explain a prediction, but they are applied after a GNN has already been trained. Here, we propose a meta-\textcolor{bs}{explainer} for improving the level of explainability of a GNN directly at training time, by steering the optimization procedure towards {\color{bs}minima that allow post-hoc explainers to achieve better results, without sacrificing the overall accuracy of the GNN}. Our framework (called MATE, MetA-Train to Explain) jointly trains a model to solve the original task, e.g., node classification, and to provide easily processable outputs for downstream algorithms that explain the model's decisions in a human-friendly way. In particular, we meta-train the model's parameters to quickly minimize the error of an instance-level GNNExplainer trained on-the-fly on randomly sampled nodes. The final internal representation relies upon a set of features that can be `better' understood by an explanation algorithm, e.g., another instance of GNNExplainer. Our model-agnostic approach can improve the explanations produced for different GNN architectures and use any instance-based explainer to drive this process. Experiments on synthetic and real-world datasets for node and graph classification show that we can produce models that are consistently easier to explain by different algorithms. Furthermore, this increase in explainability comes at no cost {\color{bs}to} the accuracy of the model. 
\end{abstract}


\section{Introduction}
\label{sec:introduction}

\IEEEPARstart{G}{raph} neural networks (GNNs) are neural network models designed to adapt and perform inference on graph domains, i.e., sets of nodes with sparse connectivity \cite{bacciu2020gentle,wu2020comprehensive,li2020fast,spinelli2020apgcn}. While a few models were already proposed in-between 2005 and 2010 \cite{gori2005new,scarselli2008graph,micheli2009neural,gallicchio2010graph}, the interest in the literature has increased dramatically over the last few years, thanks to the broader availability of data, processing power, and automatic differentiation frameworks. This is part of a larger movement towards applying neural networks to more general types of data, such as manifolds and points clouds, going under the name of `geometric deep learning' \cite{bronstein2021geometric}.

Not surprisingly, GNNs have been applied to a very broad set of scenarios, from medicine \cite{vretinaris2021medical} to road transportation \cite{zhou2020variational}, many of which possess significant risks and challenges and a potentially large impact on the end-users. Because of this, several researchers have investigated techniques to help explain the predictions done by a trained GNN, such as identifying the most critical portions of the graph that contributed to a certain inference \cite{ying2019gnnexplainer}, to help mitigate those risks and simplify the deployment of the models. Explanation methods can be broadly categorized as model-level explainers \cite{yuan2020xgnn,luo2020parameterized,schlichtkrull2020nlp}, which try to extract global explanatory patterns from the trained model, and instance-level algorithms \cite{ying2019gnnexplainer,huang2020graphlime,lucic2021cf,faber2020contrastive, yuan2021x}, which try to explain individual predictions performed by the model. In this paper, we focus on the latter group, although we hypothesize that the ideas underlying the algorithm we propose can also be extended to the former.

More generally, a limitation of most explanatory methods is that they are applied only \textit{after} a model has been trained. However, not every {\color{bs}trained} GNN is necessarily easy to explain. In critical scenarios, the end-user has no straightforward way to possibly trade-off a small amount of accuracy to increase the quality of the explanations. We hypothesize that because of the highly non-convex shape of the optimization landscape of a neural network, multiple models can have similar accuracy but possibly different behaviours when explained. In these scenarios, the literature currently lacks an easy way to steer the optimization of a GNN towards an appropriate level of explainability.

In this paper, we investigate MATE (\textbf{M}et\textbf{A}-\textbf{T}rain to \textbf{E}xplain) an algorithm to this end that is grounded in the meta-learning literature, most notably, Model-Agnostic Meta-Learning (MAML) \cite{finn17maml}. The key idea of this paper is to find explainable GNNs, by training models that can quickly converge to good explanations when a known instance-level explainer (e.g., GNNExplainer \cite{ying2019gnnexplainer}) is applied.

\subsection{Contribution of the paper}
\textcolor{bs}{We develop a framework to train GNNs such that they can be easily \textit{explained} using any instance-level algorithm. During training, for each iteration we first optimize the model to solve an explanation task, inspired by GNNExplainer, on a random subset of nodes. Then, we meta-update the model starting from the new estimate of its parameters, backpropagating through the explanation's steps.}

\begin{figure}[!th]
    \centering
    \includegraphics[scale=0.50]{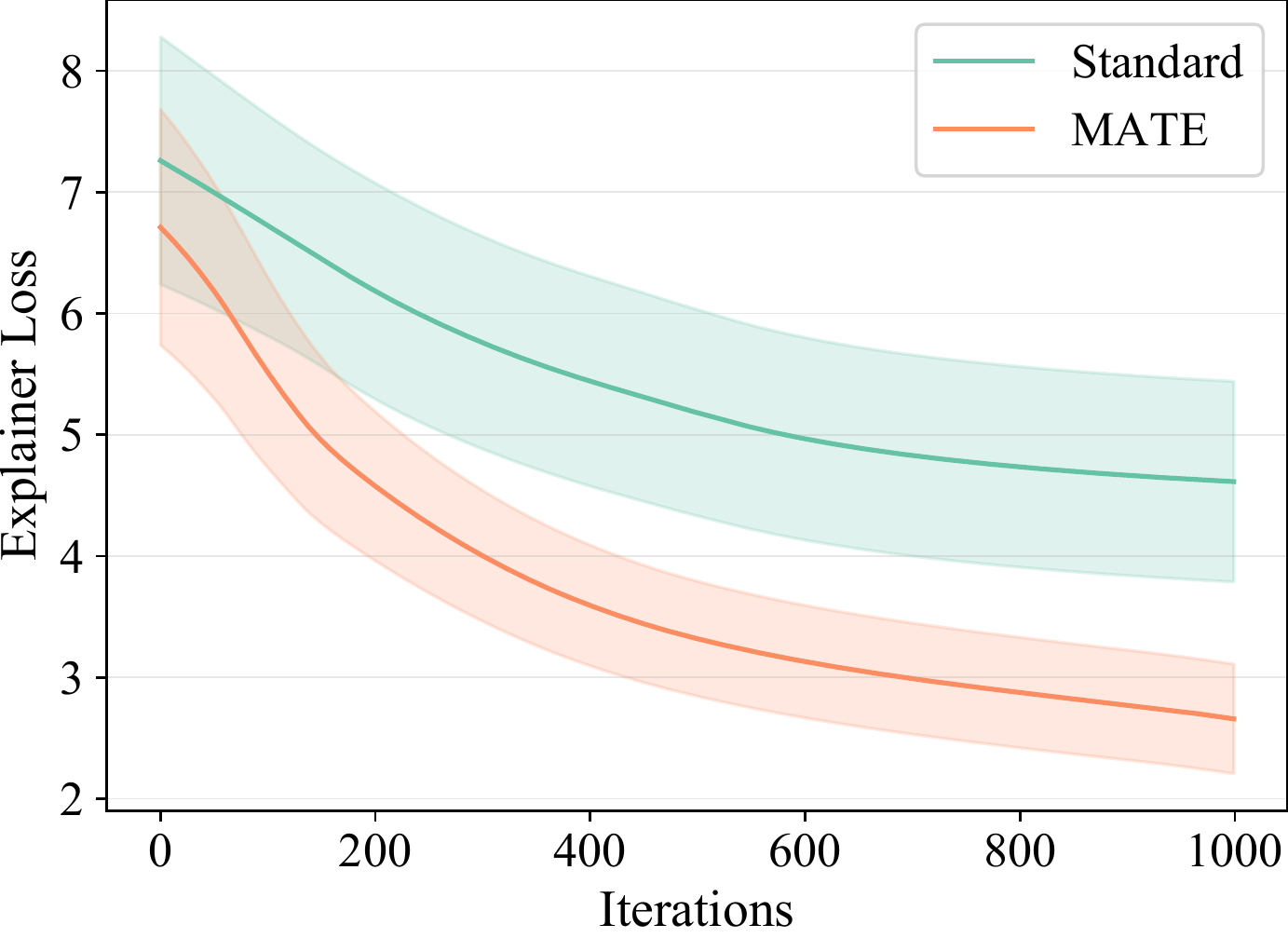}\label{fig:syn1}
    \includegraphics[scale=0.50]{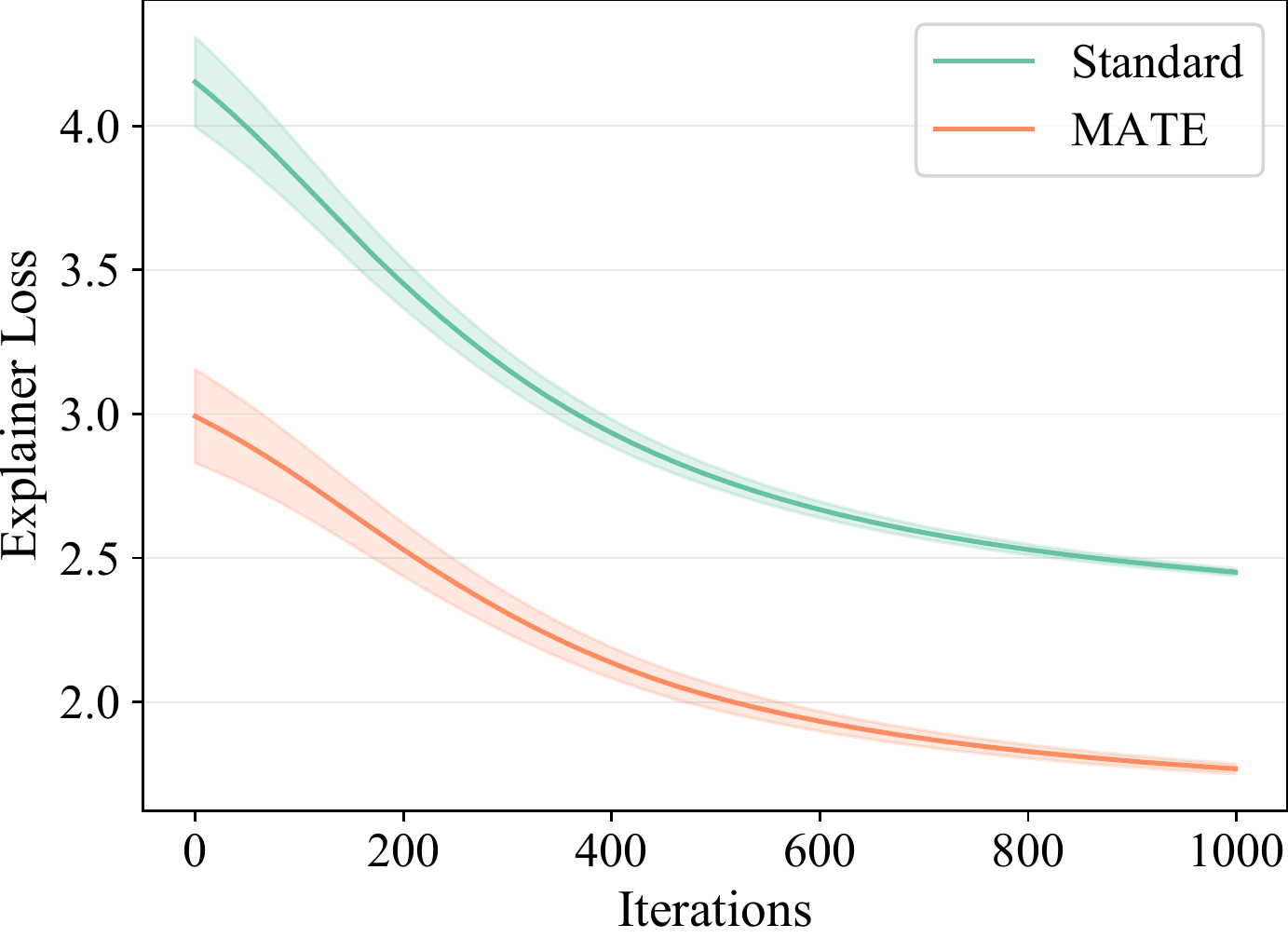}\label{fig:syn3}
    \caption{\textcolor{bs}{These plots contain GNNExplainer's losses for every node explained in the evaluation of BA-shapes (Top) and Tree-grids (Bottom) datasets. The local minima found with MATE allows the optimization process to start from a lower value. This often translates into a lower final value and sometimes steeper slopes. We believe that this is the cause of the better explainability scores obtained on MATE-trained models.
    }}
    \label{fig:intuition}
\end{figure}

\textcolor{bs}{On a wide range of experiments, we show that MATE consistently finds models whose parameters provide a better starting point for three different instance-level explainers, i.e., GNNExplainer \cite{ying2019gnnexplainer}, PGExplainer \cite{luo2020parameterized} and SubgraphX \cite{yuan2021x}. We hypothesize that MATE can efficiently steer the optimization process towards better minima (in terms of post-hoc explanation). Figure \ref{fig:intuition}, showing GNNExplainer's loss on two datasets, provides empirical support to our hypothesis. When GNNExplainer interprets the outputs of a MATE-trained model, it starts and ends its optimization process with significantly lower values. To the best of our knowledge, this is the first work proposing an algorithm to improve the degree of explainability of a GNN at training time through a meta-learning framework.}

\subsection{Organization of the paper}
The rest of the paper is structured as follows. \textcolor{bs}{In the next Section we overview the existings works in the literature.} In Section \ref{sec:preliminaries}, we introduce the framework of message-passing GNNs (Section \ref{subsec:graph_neural_networks}), instance-level explainers for GNNs (Section \ref{subsec:instance_level_explanations_gnns}), and we describe in detail GNNExplainer (Section \ref{subsec:gnnexplainer}). Then, in Section \ref{sec:proposed_approach} we introduce MATE, integrating GNNExplainer in a meta-learning bilevel optimization problem to train more interpretable networks. We evaluate our algorithm extensively in Section \ref{sec:experimental_evaluation}, where we compare the explanations we obtain when running both GNNExplainer and PGExplainer on the trained networks. Finally, we conclude with some final remarks and a series of future improvements in Section \ref{sec:conclusions_and_future_works}.

\subsection{Related Works}
\textcolor{bs}{Providing instance-level explanations in graph domains is more challenging than in other domains (such as computer vision) because of the richness of the underlying data and the general irregularity of the connectivity between nodes. In particular, a single prediction on a node of the graph can depend simultaneously on the features of the node itself, on the features of neighbouring nodes (because of the way a GNN diffuses the information over the graph), and even on graph-level properties or specific properties of the community in which the node is residing \cite{yuan2020xgnn}. Instance-level methods, such as the seminal GNNExplainer \cite{ying2019gnnexplainer}, work by extracting highly sparse masks from the computational subgraph underlying a single prediction to identify relevant node and edge features. PGExplainer \cite{luo2020parameterized} learns a parameterized model trained on the entire dataset to predict edge importance. GraphMask \cite{schlichtkrull2020nlp} follows a similar approach but predicts which edges can be dropped without changing the model's prediction. Furthermore, it computes the importance of an edge for every layer while PGExplainer only focuses on the input space. SubgraphX \cite{yuan2021x} explains its predictions by exploring different subgraphs with Monte Carlo tree search. It uses Shapley values to measure the subgraph importance and guide the search. The literature contains different types of algorithms like global explainers \cite{yuan2020xgnn}, counterfactual explainers \cite{lucic2021cf}, and algorithms developed for heterogeneous graphs \cite{yang2020ete}. For a complete review, we suggest the work of \cite{yuan2020tax}.} 

\textcolor{bs}{Our algorithm strongly differs from the literature reviewed up to this point. In fact, it is not an explainer, but rather a training procedure drawing inspiration from MAML \cite{finn17maml}, whose objective is to facilitate the work of post-hoc explainers.} MAML is a bi-level optimization method that learns the parameters of a neural network to prepare it for fast adaptation to different tasks, i.e., it finds a set of model parameters that stay relevant for several tasks rather than a single one. Formally, MAML achieves this by adapting the model's parameters $\theta$ over a randomly sampled batch of tasks using a few gradient descent updates and a few examples drawn from each one. This process generates different versions of the model's parameters, each one adapted for a specific task. Then, it exploits these modified parameters as the starting point for the meta-update of the global model's parameters. From the trained model, the model can adapt to any additional task sampled from the same distribution with only a small number of gradient descent steps. \textcolor{bs}{In our extension of MAML, we identify a task as the explanation of a given node or a graph using a specific instance-level explainer. MATE's overall goal is to modify the training procedure of the GNN such that, after training, it is easier for any explainer to find an optimum of its corresponding optimization problem. To this end, we introduce an additional set of parameters that work as a surrogate for the explainer during the GNN's training. This surrogate is trained on-the-fly and guides the inner loop optimization in which we adapt the GNN to the explanation. We hypothesize that this optimization pushes the GNN towards a set of parameters that provide better starting point for the explanation process.}

\section{Preliminaries}
\label{sec:preliminaries}

\subsection{Graph neural networks}
\label{subsec:graph_neural_networks}

The basic idea of GNNs is to combine local (node-wise) updates with suitable message passing across the graph, following the graph topology. In particular, we can represent the input graph $\mathcal{G}$ with the $n \times d$ matrix $X$ collecting all node features (with $n$ being the number of nodes and $d$ the number of features for each node) and by the adjacency matrix $A \in \left\{0, 1\right\}^{n\times n}$ encoding its topology. A two-layer Graph Convolutional Network (GCN) \cite{kipf2017semi} working on it is defined as:
\begin{equation}
    f_\theta(\mathcal{G}) = f_\theta(X,A) = \text{softmax}\left(D \phi \left (D X \theta_1\right )\theta_2 \right)\,,
    \label{eq:gcn_layer}
\end{equation}
where $\phi$ is an element-wise nonlinearity (such as the ReLU $\phi(\cdot) = \max\left(0, \cdot\right)$), $D$ is a diffusion operator like the normalized Laplacian or any appropriate shift operator defined on the graph. The model is parametrized by the vector of trainable weights $\theta = [\theta_1,\theta_2]$. The softmax function normalizes the output to a probability distribution over predicted output classes.
For tasks of node classification \cite{kipf2017semi}, we also know the desired label for a subset of nodes, and we wish to infer the labels for the remaining nodes. Graph classification is easily handled by considering sets of graphs defined as above, with a label associated with every graph, e.g., \cite{gilmer2017neural}. In this case, we need a pooling operation before the softmax to compress the node representations into a global representation for the entire graph. In both scenarios, we optimize the network with a gradient-based optimization, minimizing the cross-entropy loss:
\begin{equation}
    \mathcal{L} =  - \sum_{c=1}^{C} 1[y=c] \text{log} P_\theta(Y=y \vert\; \mathcal{G})\,,
    \label{eq:loss}
\end{equation}
where $C$ is the total number of classes, $1$ is the indicator function and $P_\theta(Y=y \vert\; \mathcal{G})$ is the probability assigned by the model $f_\theta$ to class $y$ for a single node or graph, depending on the task we wish to solve.

\subsection{Instance-level explanations for GNNs}
\label{subsec:instance_level_explanations_gnns}
Instance-level methods provide input dependant explanations, by identifying the important input features for the model's predictions. There exist different strategies for extracting this information. Following the taxonomy introduced in \cite{yuan2020tax}, we focus our analysis on perturbation-based methods \cite{ying2019gnnexplainer,luo2020parameterized,lucic2021cf,schlichtkrull2020nlp}. All these algorithms use a common scheme. They monitor the prediction's change with different input perturbations to study the importance scores associated with each edge or node's features. These methods generate some masks associated with the graph features. Then the masks, treated as optimization parameters, are applied to the input graph to generate a new graph highlighting the most relevant connections. This new graph is fed to the GNN to evaluate the masks and update them following different rules. The important features for the predictions should be the ones selected by the masks.

Following the notation from the previous sub-section, let $f_\theta(\mathcal{G})$ be a trained GNN model parametrized by weights $\theta$, making predictions from the input graph $\mathcal{G}$ having node features $X$ and the adjacency matrix $A$. Given a node $v$ whose prediction we wish to explain, we denote by $\mathcal{G}_v^c = (X_v^c, A_v^c)$ the subgraph participating in the computation of $f_\theta(\mathcal{G})$. For example, for a two-layer GCN, $\mathcal{G}_v^c$ contains all the neighboors up to order two described with the associated adjacency matrix $A_v^c \in \{0,1\}^{n \times n}$ and their set of features $X_v^c= \{ x_j | j \in X_v^c \}$.
Instance-level explanations methods generate random masks for the input graph $\mathcal{G}_v^c$ and treat them as training variables $W=[M_v^x, M_v^a]$. These masks are applied to the computational subgraph generating the explanation subgraph $\mathcal{G}_v^e = (X_v^e, A_v^e)$. The explanation is the solution of the optimization problem over $\mathcal{G}_v^e$:
\begin{equation}
    G_v^{e*} =  \underset{W}{\arg\min}\left\{ \mathcal{L}_e(\mathcal{G}_v^c, \theta, W) \right\} \,,
    \label{eq:explainable_loss}
\end{equation}
where $\mathcal{L}_e$ is an explanation objective, such that lower values correspond to `better' explanations. In the following, we describe briefly the procedure for GNNExplainer \cite{ying2019gnnexplainer}, although our proposed approach can be easily extended to any method of the form \eqref{eq:explainable_loss}.
\subsection{GNNExplainer}
\label{subsec:gnnexplainer}

In GNNExplainer \cite{ying2019gnnexplainer} the masks are applied to the computational subgraph via pairwise multiplication:
\begin{equation}
\mathcal{G}_v^e = (X_v^e, A_v^e) = \left(X_v^c \odot \sigma(M_v^x), A_v^c \odot \sigma(M_v^a)\right)\,,
\end{equation}
where $W = [M_v^x,M_v^a]$ are the explainer's parameters, $\odot$ denotes the element-wise multiplication, and $\sigma$ denotes the sigmoid.
Then, it optimizes the masks by maximizing the mutual information (MI) between the original prediction and the one obtained with the masked graph. Different regularization terms encourage the masks to be discrete and sparse. Formally, GNNExplainer defines the following optimization framework:
\begin{equation}
    \underset{\mathcal{G}_v^e}{\arg\max} \;\; \text{MI}(Y, \mathcal{G}_v^e) = H(Y \vert\; \mathcal{G}_v^c) - H(Y \vert\; \mathcal{G}_v^e)\,,
\end{equation}
where MI quantifies the change in the conditional entropy $H(\cdot)$ (or probability prediction) when $v$'s computational graph is limited to the explanation subgraph. The first term of the equation is constant for a trained GNN. Hence, maximizing the mutual information corresponds to minimizing the conditional entropy:
\begin{equation}
    H(Y| \mathcal{G}_v^e) = - \mathbb{E}_{Y|\mathcal{G}_v^e} [\text{log} P_\theta \left( Y | \mathcal{G}_v^e \right)]\,,
\end{equation}
which gives us the subgraph that minimizes the uncertainty of the network's (parametrized by $\theta$) prediction when the GNN computation is limited to $\mathcal{G}_v^e$.

When we are interested in the reason behind the prediction of a certain class for a certain node the conditional entropy is replaced with a cross-entropy objective. Finally, we can use gradient descent-based optimization to find the optimal values for the masks minimizing the following objective:
\begin{equation}
    \mathcal{L}_e = - \sum_{c=1}^{C} 1[y=c] \text{log} P_\theta \left(Y=y \vert\; \mathcal{G}_v^e \right) \,.
    \label{eq:exp_loss}
\end{equation}

GNNExplainer deploys some regularization strategies to obtain its explanations. Firstly, an element-wise entropy encourages the masks to be discrete. Secondly, the sum of all masks elements (equivalent to their $\ell_1$ norm) penalizes large subgraphs. All these strategies contribute to the fact that $\mathcal{G}_v^e$ tends to be a small connected network containing the node to be explained.

\section{Proposed approach}
\label{sec:proposed_approach}

\begin{figure}
    \centering
    \includegraphics[width=0.9\columnwidth]{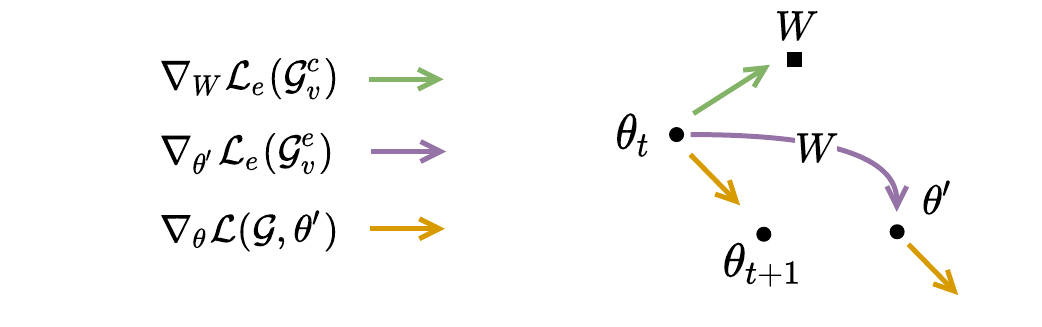}
    \caption{\textcolor{bs}{Diagram of our algorithm, which optimizes for an explainable representation $\theta$.}}
    \label{img:gd}
\end{figure}

The methods described in Section \ref{subsec:instance_level_explanations_gnns} work by explaining a single instance after the GNN has been trained, decoupling the two steps. In this section, we aim to train models that can be biased towards producing explainable predictions during their training. We first define the problem setup and present the general form of our algorithm.
MAML \cite{finn17maml} works by optimizing models that can be quickly trained on new tasks. In our framework, we define an `explanation task' on a randomly sampled node and train a model that can quickly converge to a good explanation according to the desired metric (in our case, GNNExplainer).
Our goal is a model with more interpretable outputs. By adapting the GNN's parameters for these `explanation tasks', we promote an internal representation based on easily interpretable features (see Fig. \ref{img:gd}).

\subsection{Problem Setup}

We define the main task, being node or graph classification, as optimizing the original objective function $\mathcal{L}$ with gradient-based techniques (e.g., cross-entropy over the labelled nodes of the graph). We will take into consideration node classification tasks being the extension to graph classification trivial. Then we define a single `explanation task' \textcolor{bs}{$\mathcal{T}_v^e = \left\{f_\theta, \mathcal{G}_v^{e}, \mathcal{L}_e \right\}$} for any randomly sampled node $v$ from the graph $\mathcal{G}$. The explanation task requires \textcolor{bs}{an explanatory subgraph $\mathcal{G}_v^{e}$ produced by an explainer trained on-the-fly during the optimization of the model $f_\theta$ using its current state $\theta$. The loss $\mathcal{L}_e$ provides task-specific feedback. It's the same loss used during the explainer's optimization with $\mathcal{G}_v^{e}$ fixed instead of $\theta$.} By optimizing the model's parameters $\theta$ for a few steps of gradient descent, we will adapt the model's parameters to the explanation, producing a new set of parameters $\theta'$.
In the next Section, we will explain step by step how we use these adapted parameters to perform a meta-optimization for the model's principal objective.

\subsection{Meta-Explanation}

\begin{figure*}
    \centering
    \includegraphics[width=1.78\columnwidth]{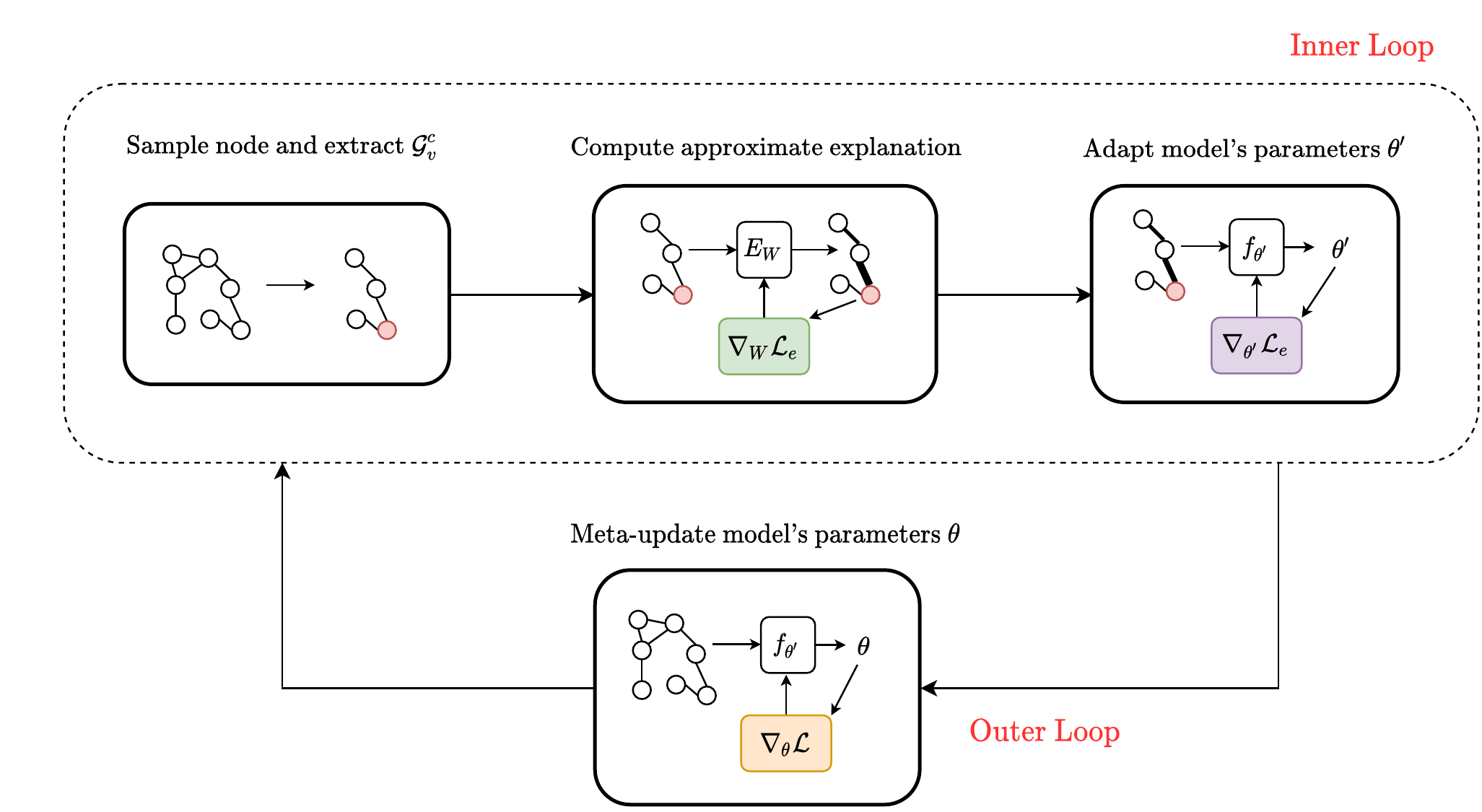}
    \caption{Schematics of our meta-learning framework for improving GNN's explainability at training time. MATE steers the optimization procedure towards more interpretable minima in the inner loop, meanwhile optimizing for the original task in the outer one. The inner loop adapts the model's parameters to a single `explanation task'. It starts with the sampling of a random node and its computational subgraph. Then, we train GNNExplainer to explain the current model's prediction. Afterwards, we can adapt the model's parameters to the `explanation task' ending in a new model's state. Finally, we meta-update the original parameters minimizing the cross-entropy loss computed with the adapted parameters.}
    \label{img:scheme}
\end{figure*}

We want to steer the optimization process to find a set of parameters that help post-hoc explanation algorithms to provide relevant interpretation of the model prediction. We present a graphical representation of our framework in Fig. \ref{img:scheme}.

\begin{algorithm}[t]
\caption{\textbf{: MATE}}
\label{alg:1}
\vspace{.1cm}
\textbf{Data:} Input Graph $\mathcal{G} = (X, A)$\\
\textbf{Require:} $\alpha, \beta, \gamma$ step size hyperparameters, (K,T) number of gradient-based optimization steps. 

\begin{algorithmic}[1]
\STATE Initialize model's parameters $\theta$.
\WHILE{ not done}
    \STATE Random sample $v$ from $\mathcal{G}$ and \\ extract computational graph $\mathcal{G}_v^c$
    \STATE Initialize explainer's parameters $W$
    \FOR{K steps}
    \STATE Compute explainer's parameters ($\theta$ fixed) \\ \textcolor{bs}{$ W = W - \delta \nabla_{W} \mathcal{L}_e(\mathcal{G}_v^c, W)$}
    \ENDFOR
    \FOR{T steps}
    \STATE Adapt model's parameters ($W$ fixed) \\ \textcolor{bs}{$ \theta' = \theta' - \alpha \nabla_{\theta'} \mathcal{L}_e(\mathcal{G}_v^e, \theta')$}
    \ENDFOR
    \STATE  Meta-update \\ $ \theta = \theta - \beta \nabla_{\theta} \mathcal{L}(\mathcal{G}, \theta')$
    
    \ENDWHILE
\end{algorithmic}
\end{algorithm}

Differently from MAML, MATE has two different optimization objectives. The first is the explanation objective $\mathcal{L}_e$ as described in \eqref{eq:exp_loss} and optimized in the inner loop. The second is the main objective $\mathcal{L}$ \eqref{eq:loss}, a standard cross-entropy loss for the main classification task, optimized in the outer loop. The first one uses the computational subgraph of a randomly sampled node. The second one instead exploits the entire graph.

To perform a single update, we start the inner loop by sampling at random a node $v$ from the graph and extracting its computational subgraph $\mathcal{G}_v^c$. Explaining $v$'s prediction will be our target or our `explanation task'. We continue by initializing and training a GNNExplainer minimizing \eqref{eq:exp_loss} for $K$ gradient steps (with $K$ being a hyper-parameter) to obtain the explanation subgraph for $v$ based on the current GNN parameters. A single update is in the form:
\begin{equation}
\textcolor{bs}{
    W = W - \delta \nabla_{W} \mathcal{L}_e(\mathcal{G}_v^c, W)\,,
}
\end{equation}
where we take the gradient of the explanation loss with respect to the explainer's parameters regarding the model's ones fixed. The step size $\delta$, like all the next step sizes, may be fixed or meta-learned.

At this point we can define our `explaination task' \textcolor{bs}{$\mathcal{T}_v^e = \left\{f_\theta, \mathcal{G}_v^{e}, \mathcal{L}_e \right\}$}. We adapt the model parameters to  $\mathcal{T}_v^e$ using $T$ gradient descent updates.
Again, a single update takes the form:
\begin{equation}
\textcolor{bs}{
    \theta' = \theta' - \alpha \nabla_{\theta'} \mathcal{L}_e(\mathcal{G}_v^e, \theta')\,,
}
\end{equation}
where $\theta'$ is the vector of the adapted parameters \textcolor{bs}{ and $\mathcal{G}_v^{e}$ is the explanation subgraph.}
We compute the gradient with respect to the model's parameters leaving the explainer's masks fixed. Like the previous update, we have another hyperparameter paired with $T$, $\alpha$, representing the step size for the adaptation.

The model's parameters are trained by optimizing for the performance of $f_{\theta'}$  with respect to $\theta$ for the main classification task, exploiting the entire graph structure. The meta-update is defined as:
\begin{equation}
\textcolor{bs}{
\theta = \theta - \beta \nabla_{\theta} \mathcal{L}(\mathcal{G}, \theta')\,,
}
\end{equation}
where $\beta$ is the meta step size.
The meta-optimization updates $\theta$ using the objective computed with the adapted model's parameters $\theta'$. We outline the framework in Algorithm \ref{alg:1}.

The meta-gradient update involves a gradient through a gradient. We use the Higher library \cite{grefenstette2019generalized} to handle the additional backward passes and to deploy the Adam optimizer \cite{kingma2014adam} to perform the actual updates.
The extension to the graph classification task is trivial. Instead of sampling a random node for the explanation, we select an entire graph from the current batch used for the model's update.

\section{Experimental Evaluation}

\label{sec:experimental_evaluation}

In this Section, we evaluate MATE with several experiments over synthetic and real-world datasets. We first describe the datasets and experimental setup. Then, we present the results on both node and graph classification. With qualitative and quantitative evaluations, we demonstrate that GNNExplainer, PGExplainer \textcolor{bs}{and SubgraphX} provide better explanations results when used on models trained with MATE, in some cases improving the state-of-the-art in explaining node/graph classification predictions. At the same time, we show that our framework does not impact the classification accuracy of the model. We based our implementation\footnote{https://github.com/ispamm/MATE} upon the code develop in \cite{holdijk2021re}.

\begin{table*}[t]
\centering
\normalsize
\caption{Visualization of the explanation subgraphs for the node classification task. Node colors represent node labels. Darkness of the edges signals importance for the classification. The ground truth motif is presented in the first row.}
\begin{tabular}{lcccccc}
  & BA-shapes & BA-community & Tree-cycles & Tree-grids \\
  \midrule
\raisebox{3\height}{Motif} & \raisebox{0.1\height}{\includegraphics[scale=0.75]{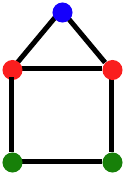}} & \raisebox{-0.1\height}{\includegraphics[scale=0.5]{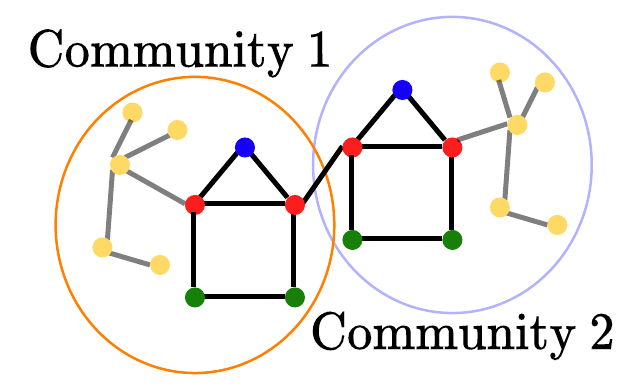}} & \raisebox{0.3\height}{\includegraphics[scale=0.75]{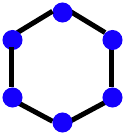}} & \raisebox{0.35\height}{\includegraphics[scale=0.75]{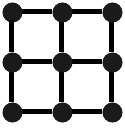}} \\
    \midrule
\raisebox{3\height}{GNNExp} & \includegraphics[scale=0.23]{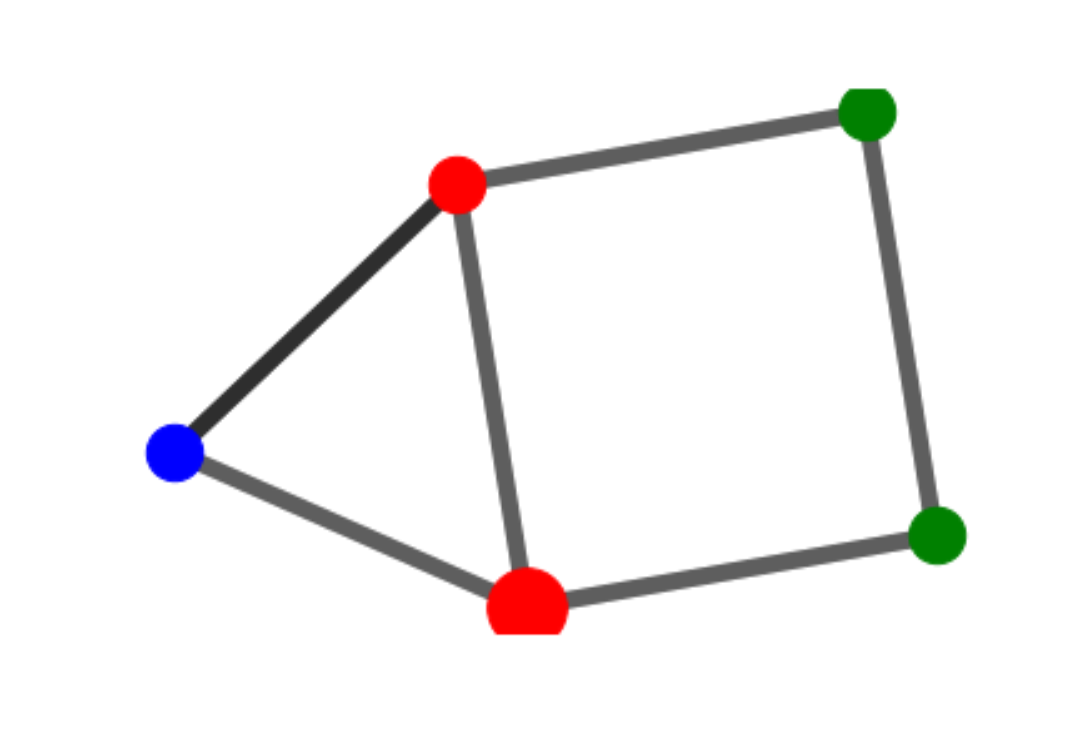} & \includegraphics[scale=0.23]{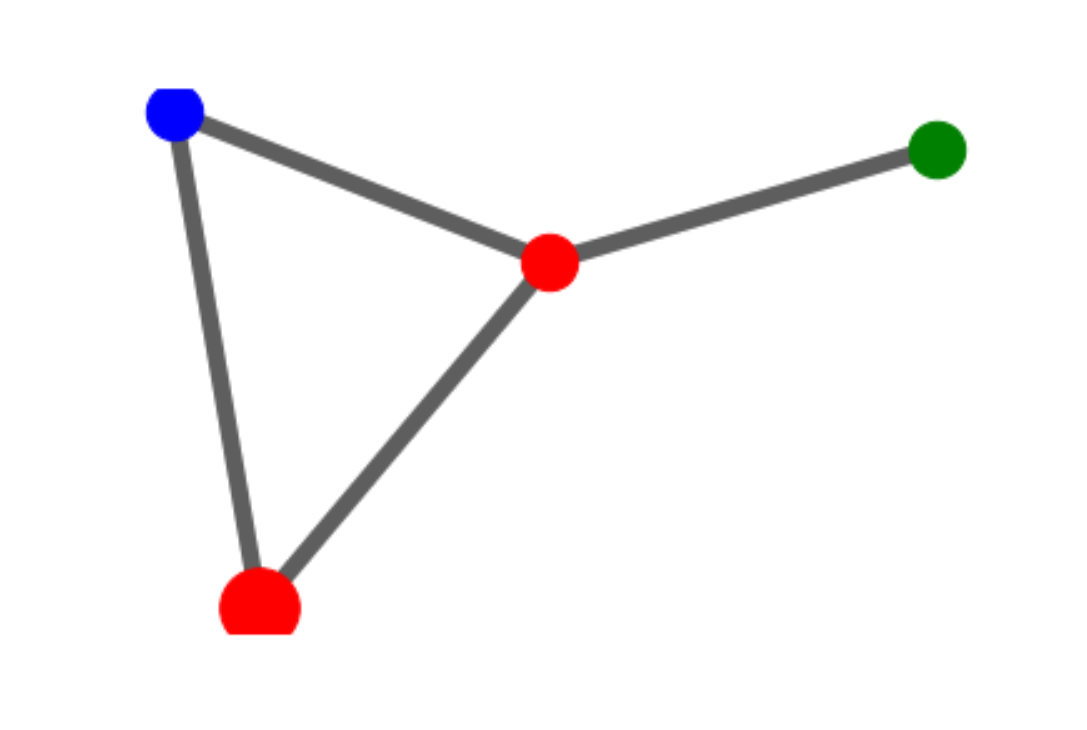} & \includegraphics[scale=0.23]{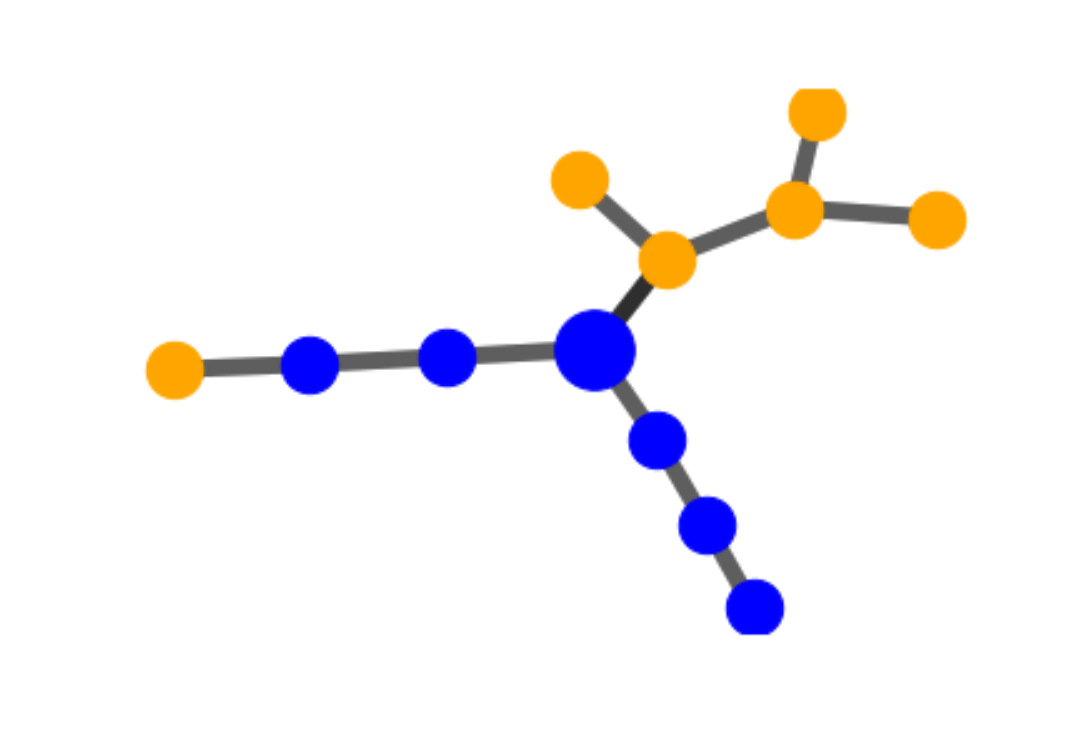} & \includegraphics[scale=0.23]{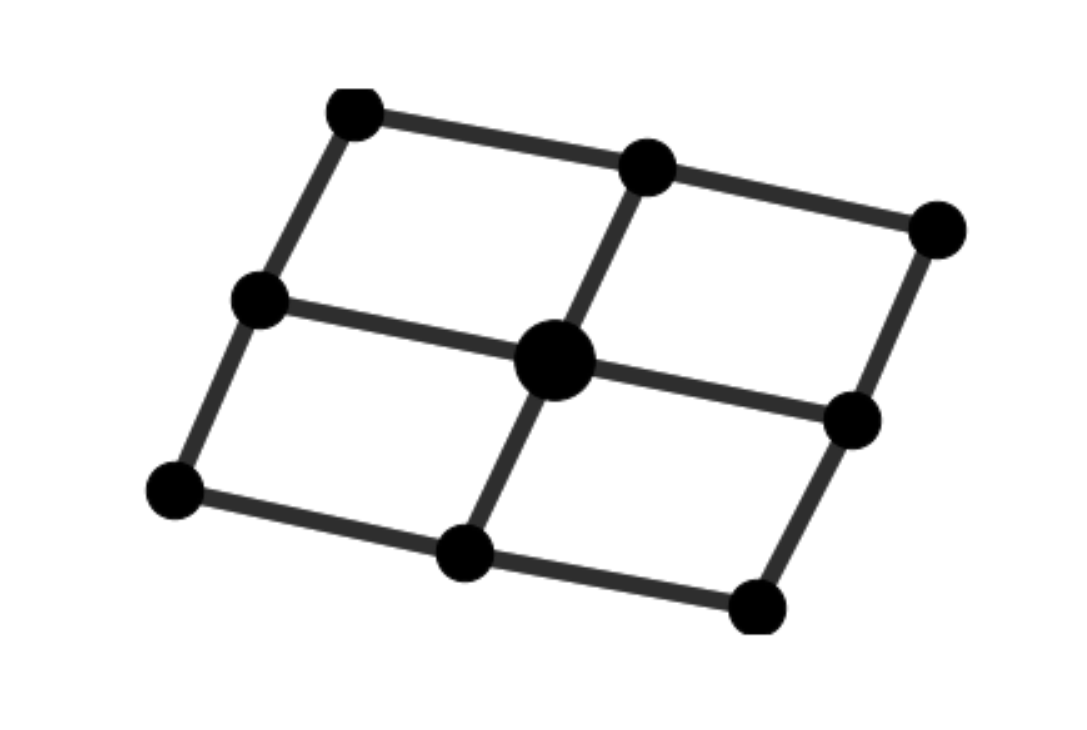} \\

\raisebox{3\height}{MATE+GNNExp} & \includegraphics[scale=0.23]{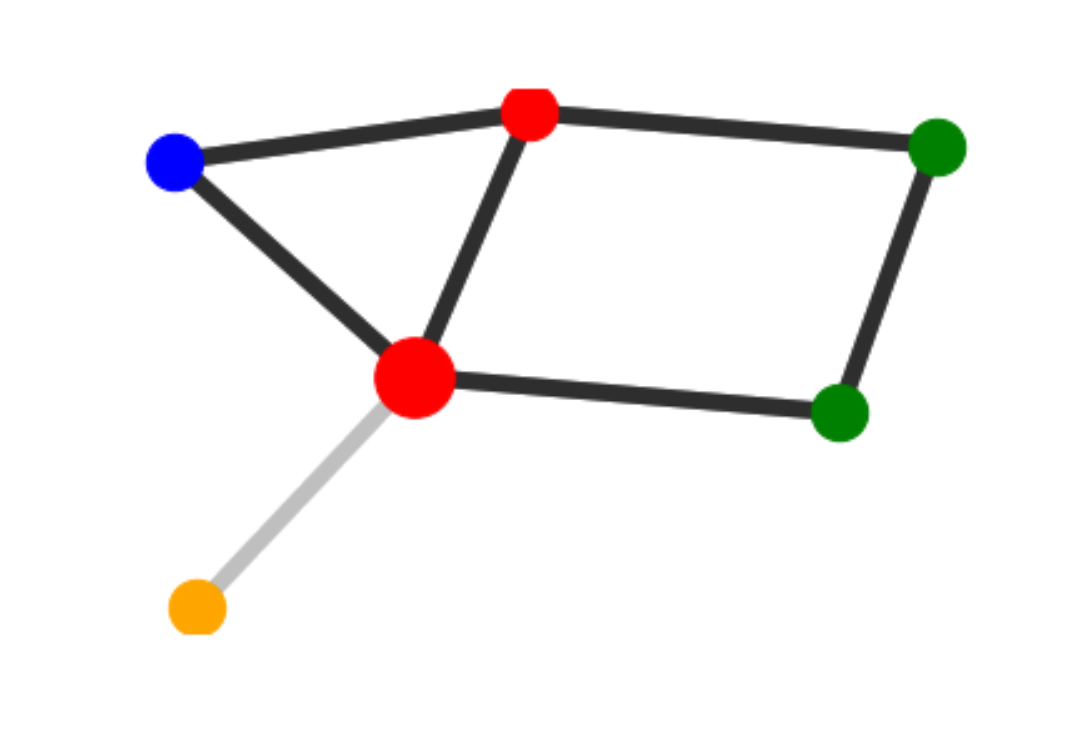} & \includegraphics[scale=0.23]{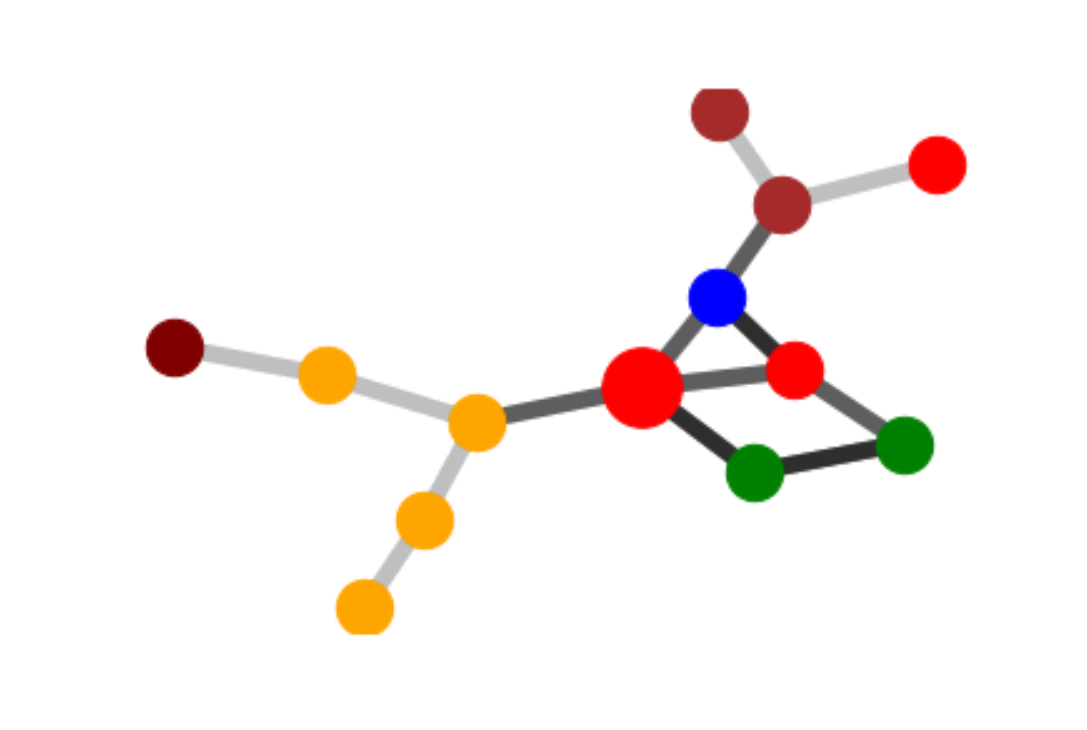} & \includegraphics[scale=0.23]{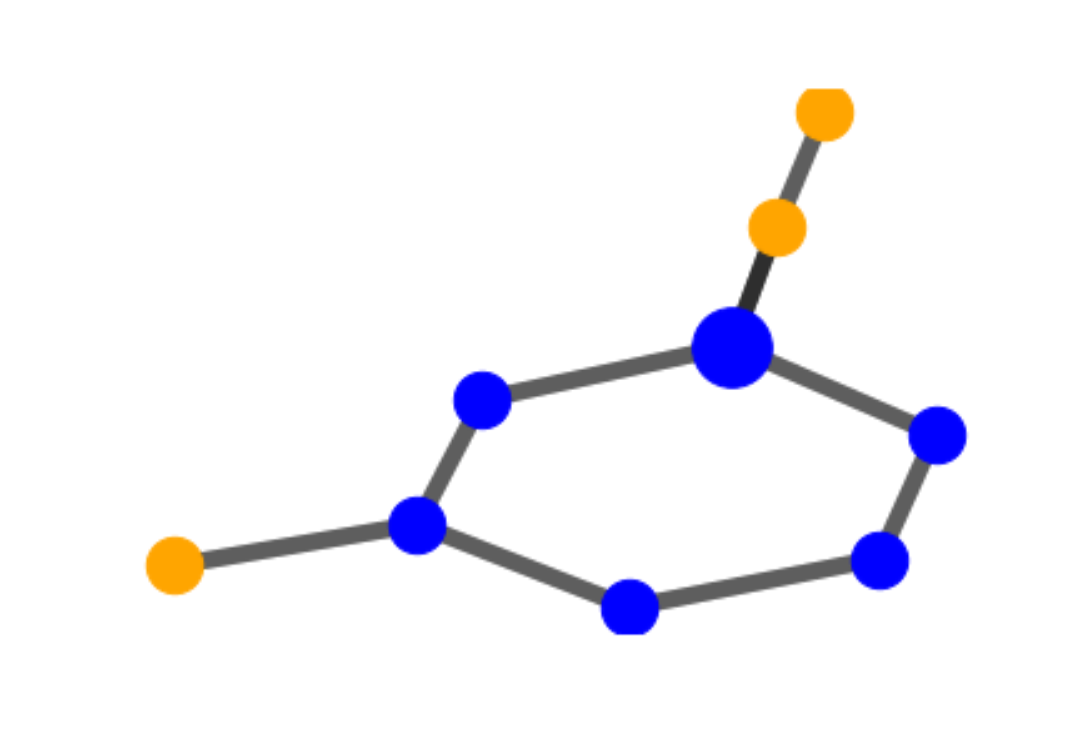} & \includegraphics[scale=0.23]{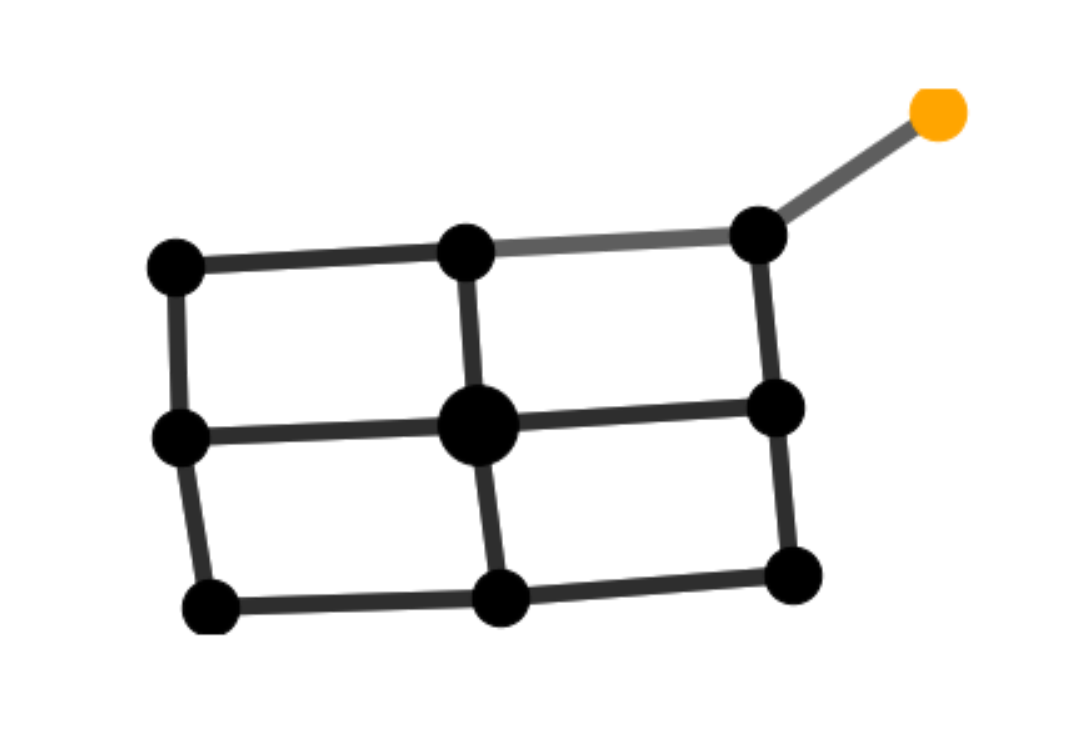}\\
\midrule

\raisebox{3\height}{PGExp} & \includegraphics[scale=0.23]{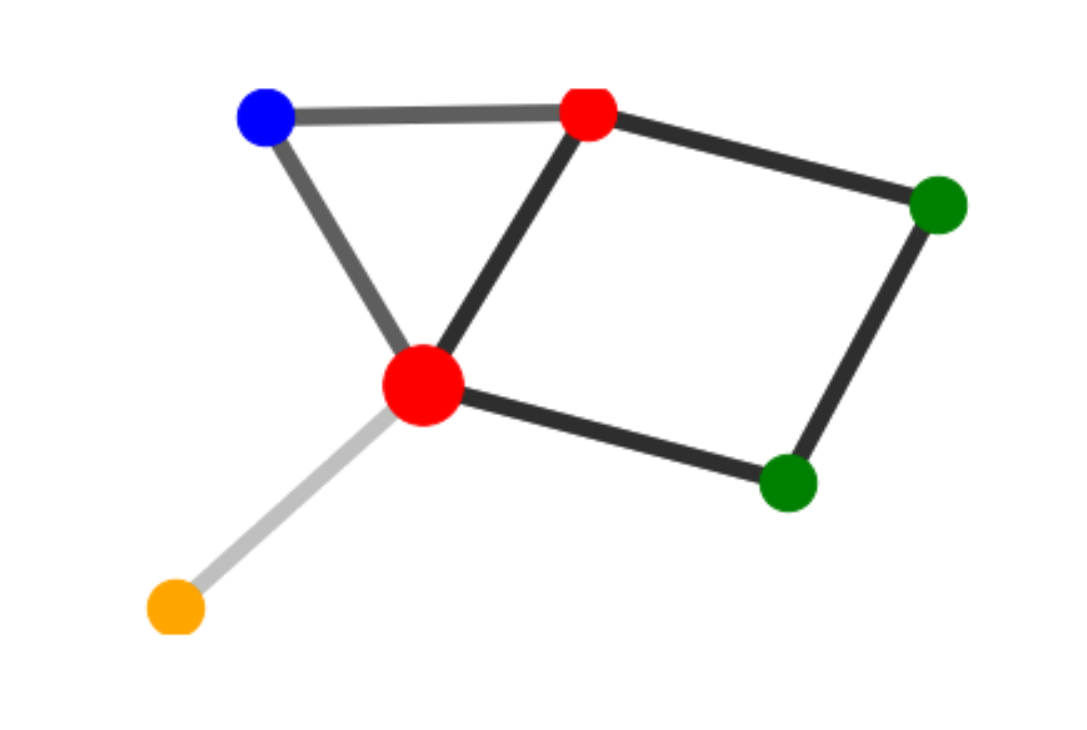} & \includegraphics[scale=0.23]{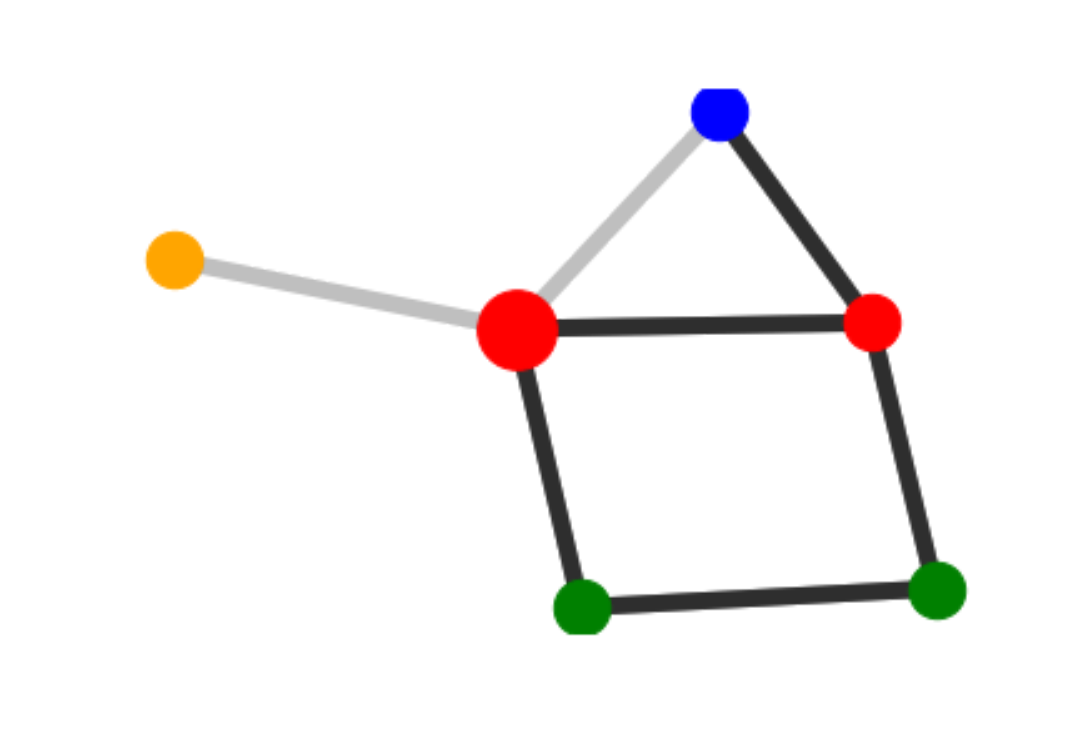} & \includegraphics[scale=0.23]{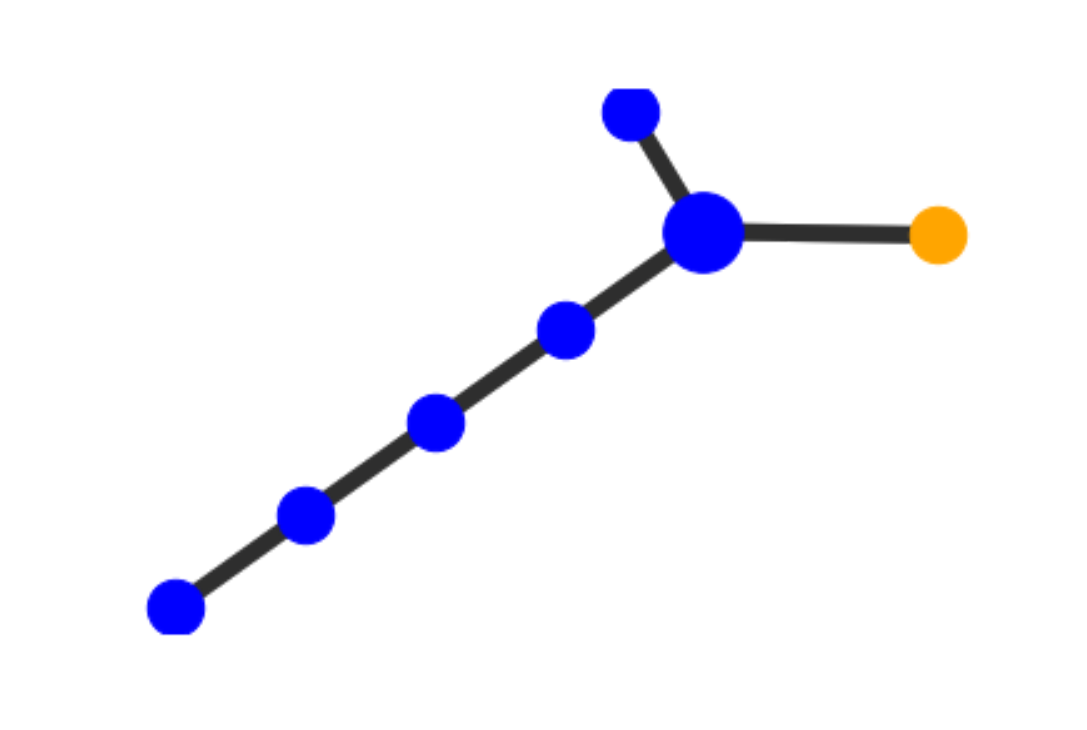} & \includegraphics[scale=0.23]{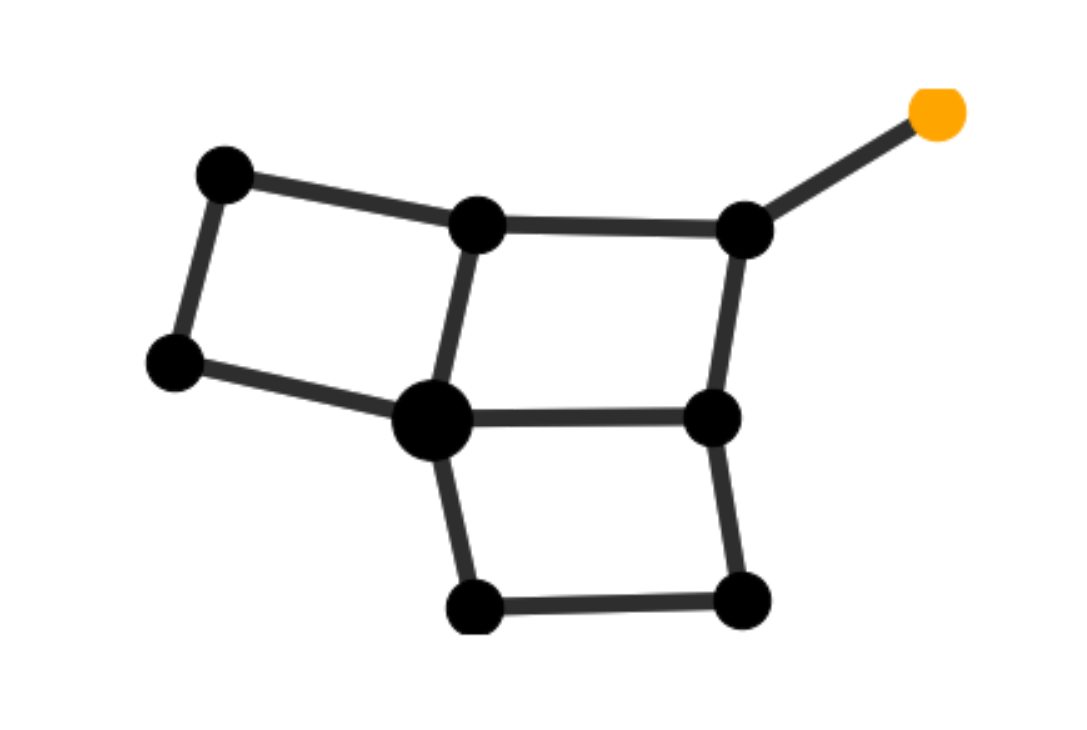} \\
\raisebox{3\height}{MATE+PGExp} & \includegraphics[scale=0.23]{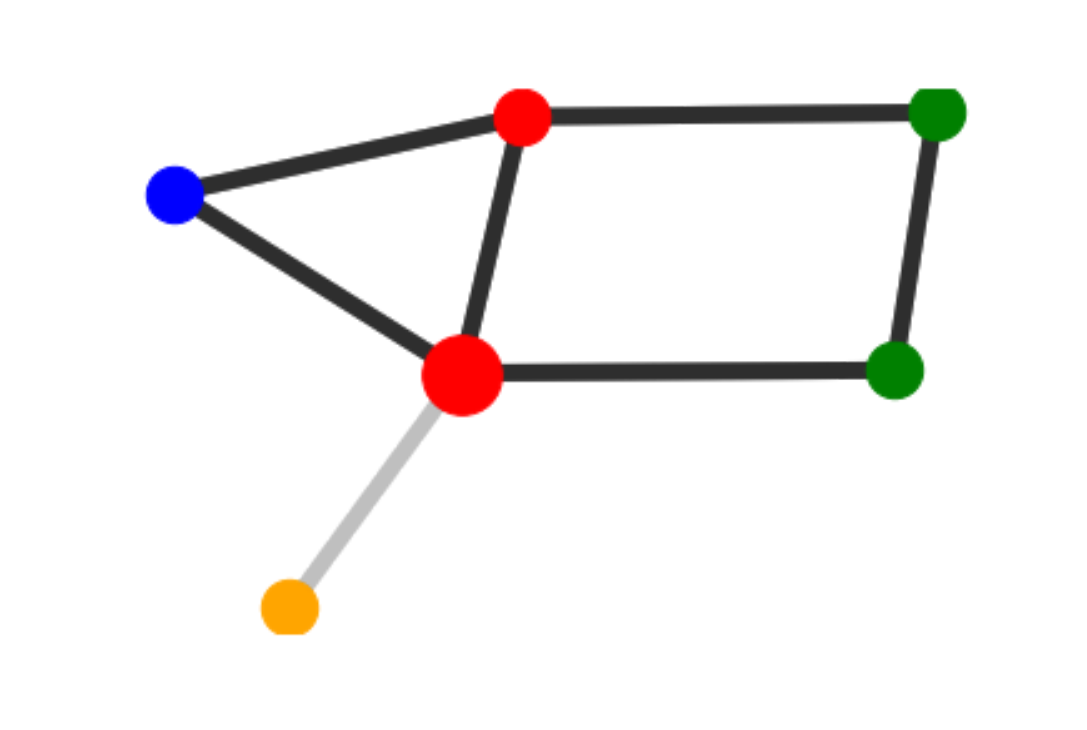} & \includegraphics[scale=0.23]{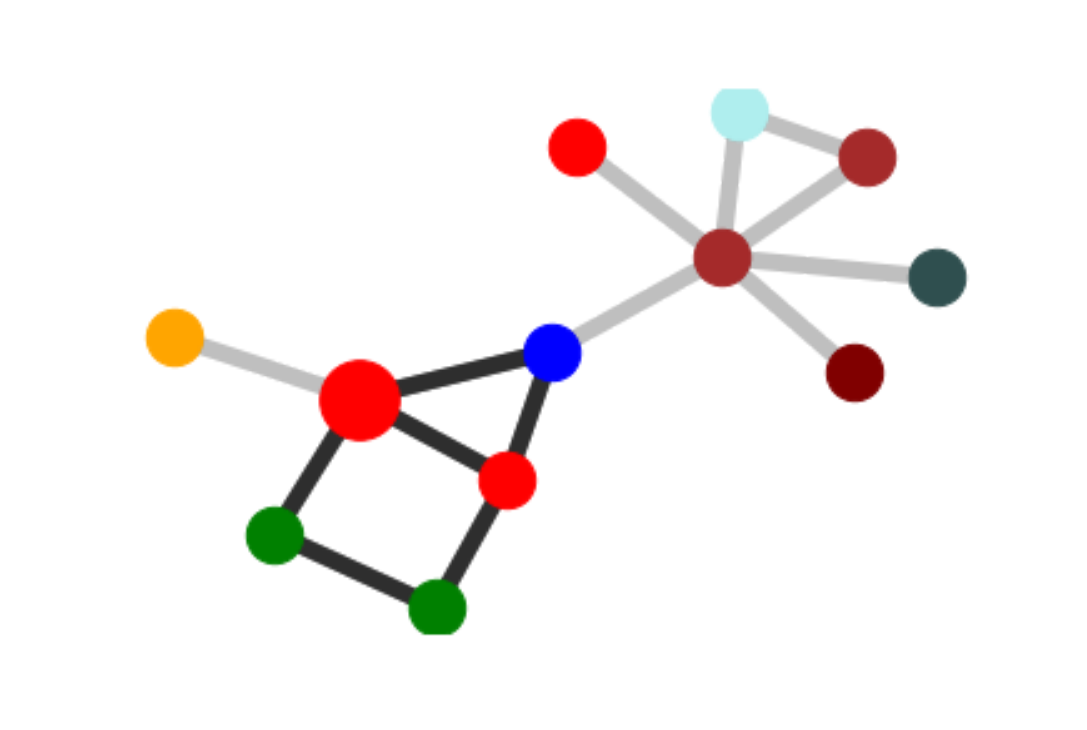} & \includegraphics[scale=0.23]{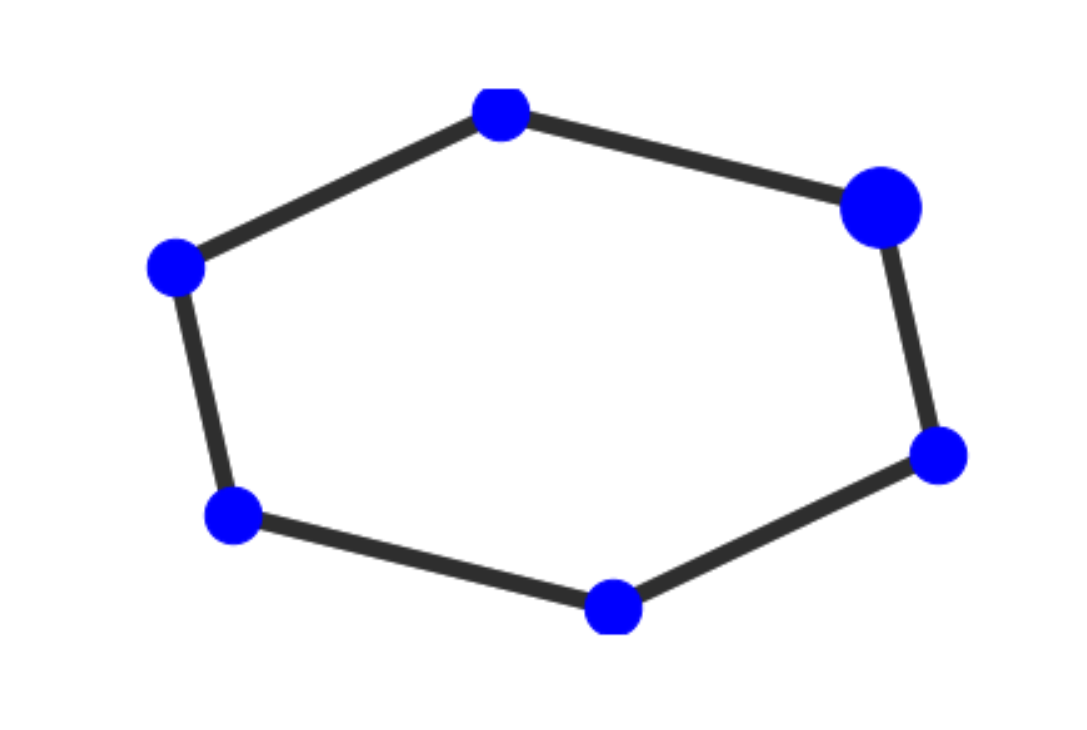} & \includegraphics[scale=0.23]{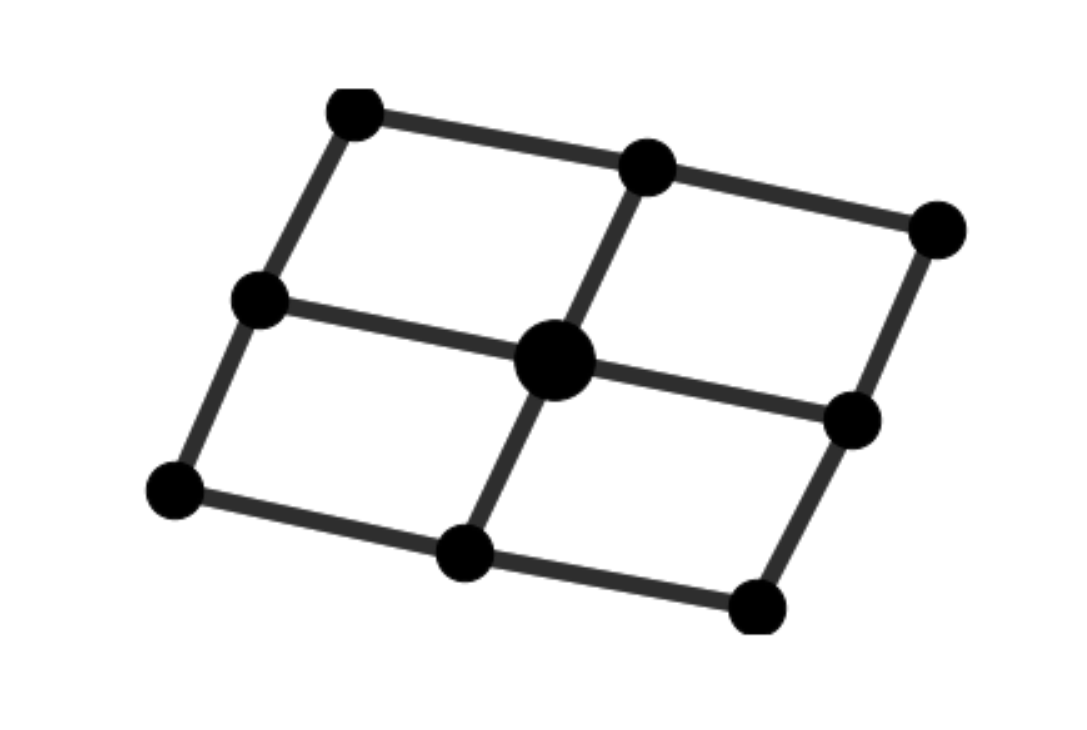} \\
\midrule

\raisebox{3\height}{SubgraphX} & \includegraphics[scale=0.10]{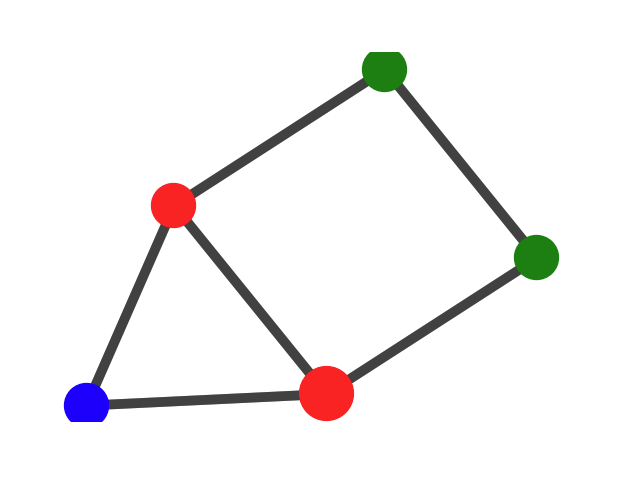} & \includegraphics[scale=0.15]{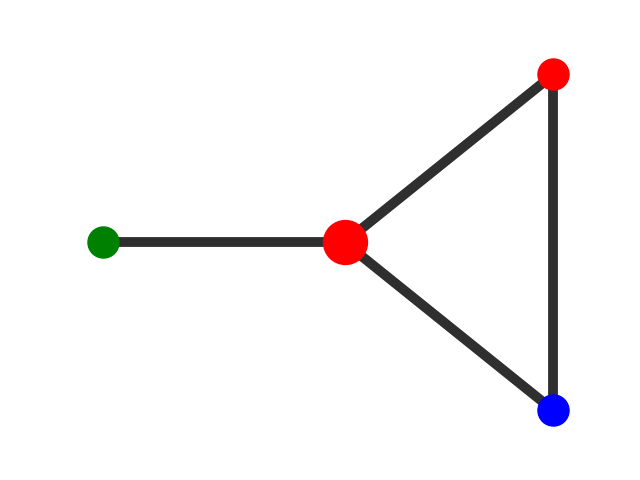} & \includegraphics[scale=0.15]{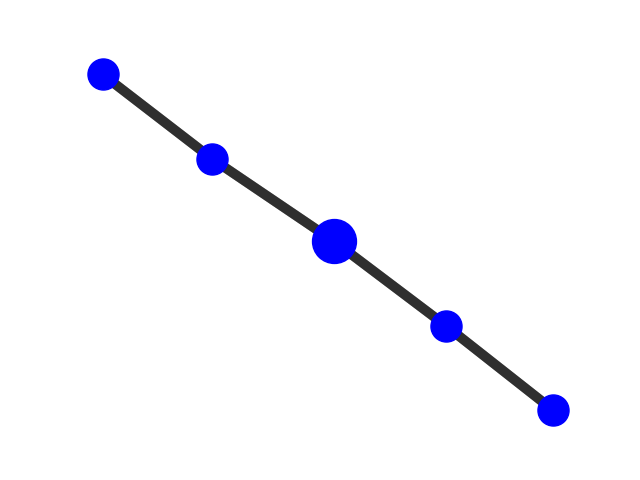} & \includegraphics[scale=0.15]{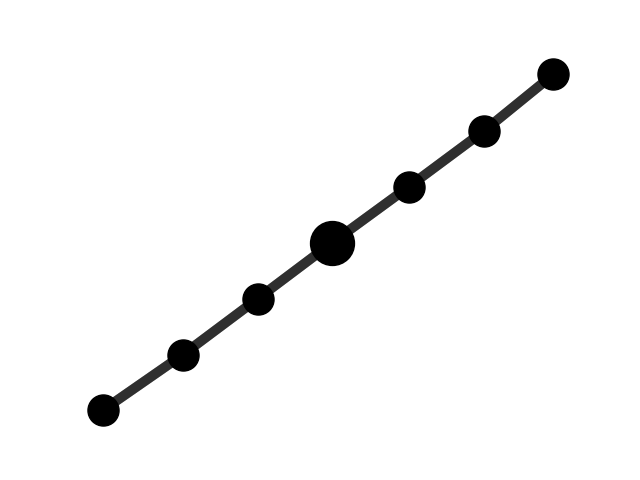} \\

\raisebox{3\height}{MATE+SubgraphX} & \includegraphics[scale=0.10]{images/bella.png} & \includegraphics[scale=0.10]{images/bella.png} & \includegraphics[scale=0.15]{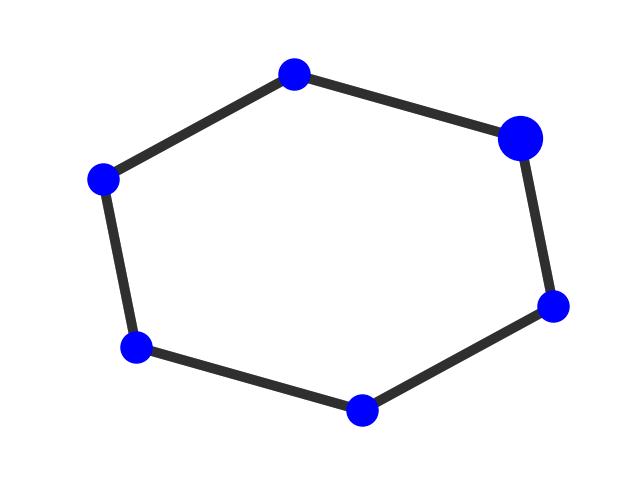} & \includegraphics[scale=0.15]{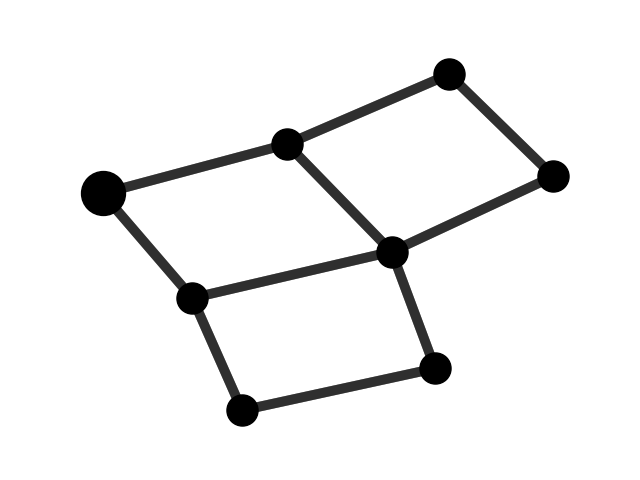} \\
\midrule

\end{tabular}
\label{tab:figPP}
\end{table*}

\begin{table}[t]
\centering
\normalsize
\caption{Visualization of the explanation subgraphs for the graph classification. Darkness of the edges signals importance for the classification. The ground truth motif is in the first row.}
\begin{tabular}{lcc}
    & BA-2motifs & MUTAG \\
    \midrule
\raisebox{2\height}{Motif} & \includegraphics[scale=0.75]{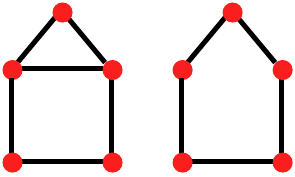} & \raisebox{-0.2\height}{\includegraphics[scale=0.75]{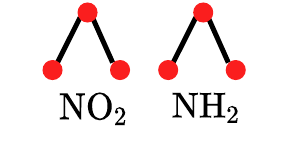}} \\
    \midrule
\raisebox{3\height}{GNNExp} & \includegraphics[scale=0.23]{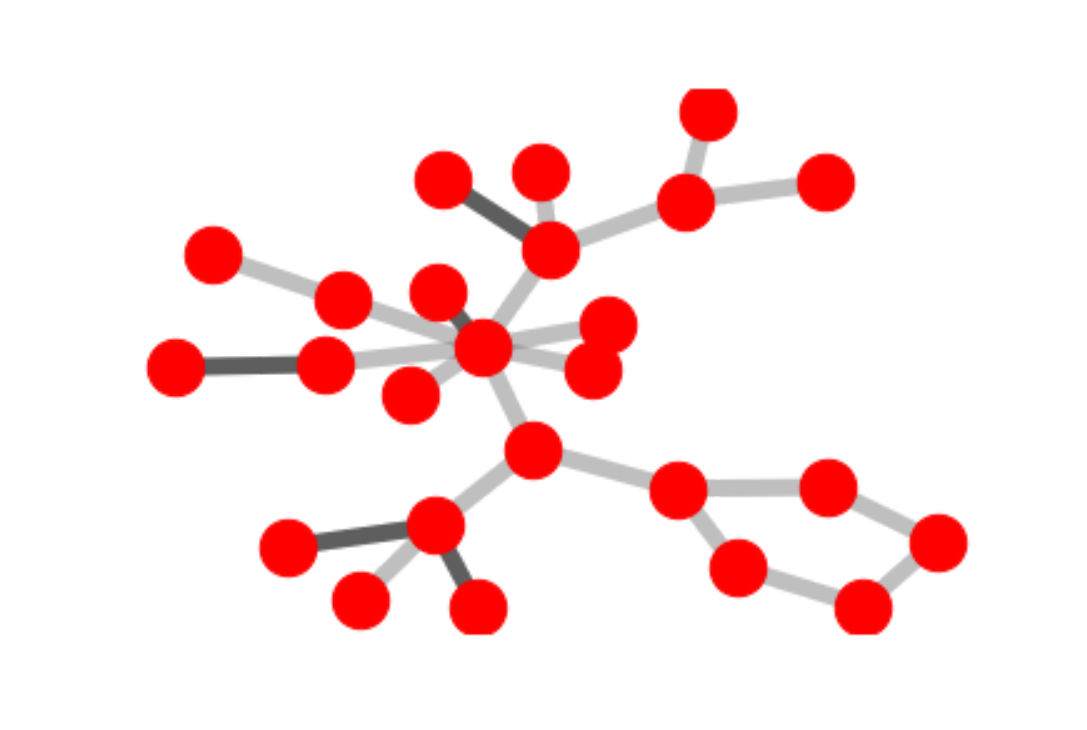} & \includegraphics[scale=0.23]{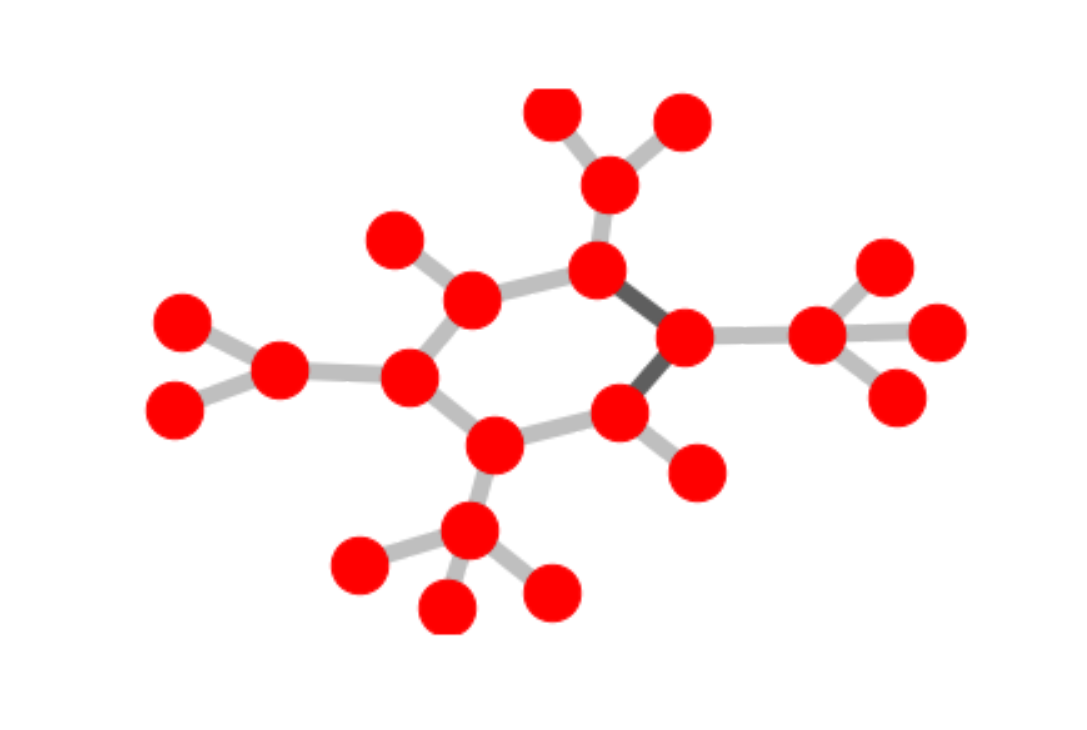} \\
\raisebox{3\height}{MATE+GNNExp} & \includegraphics[scale=0.23]{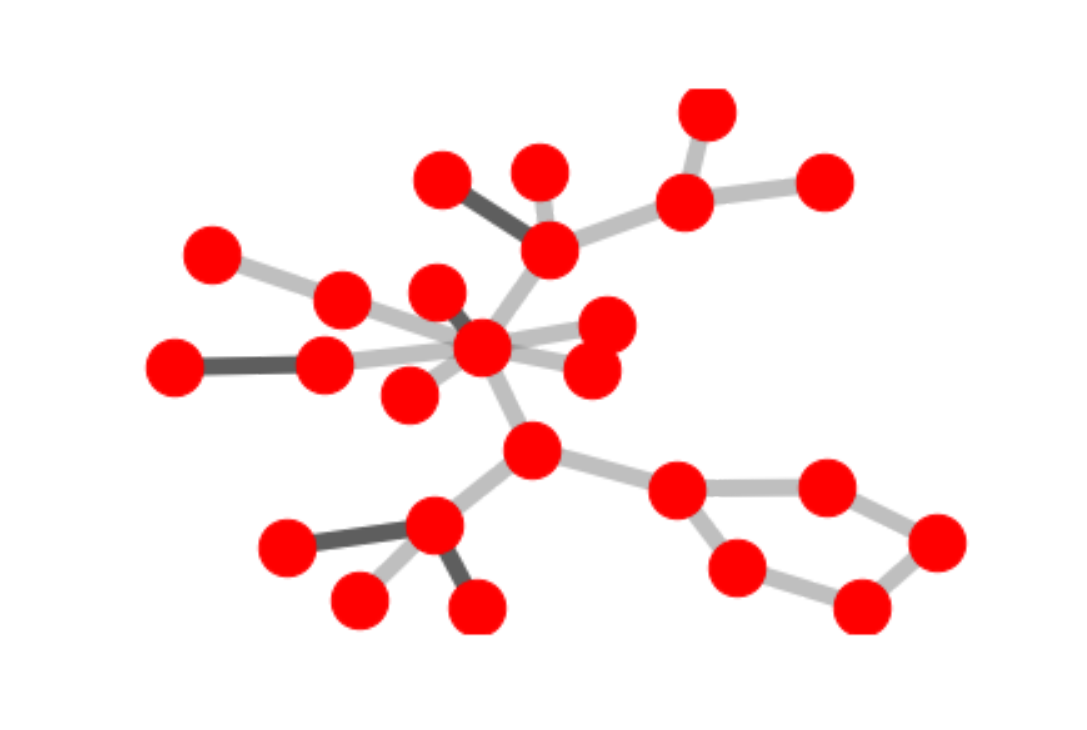} & \includegraphics[scale=0.23]{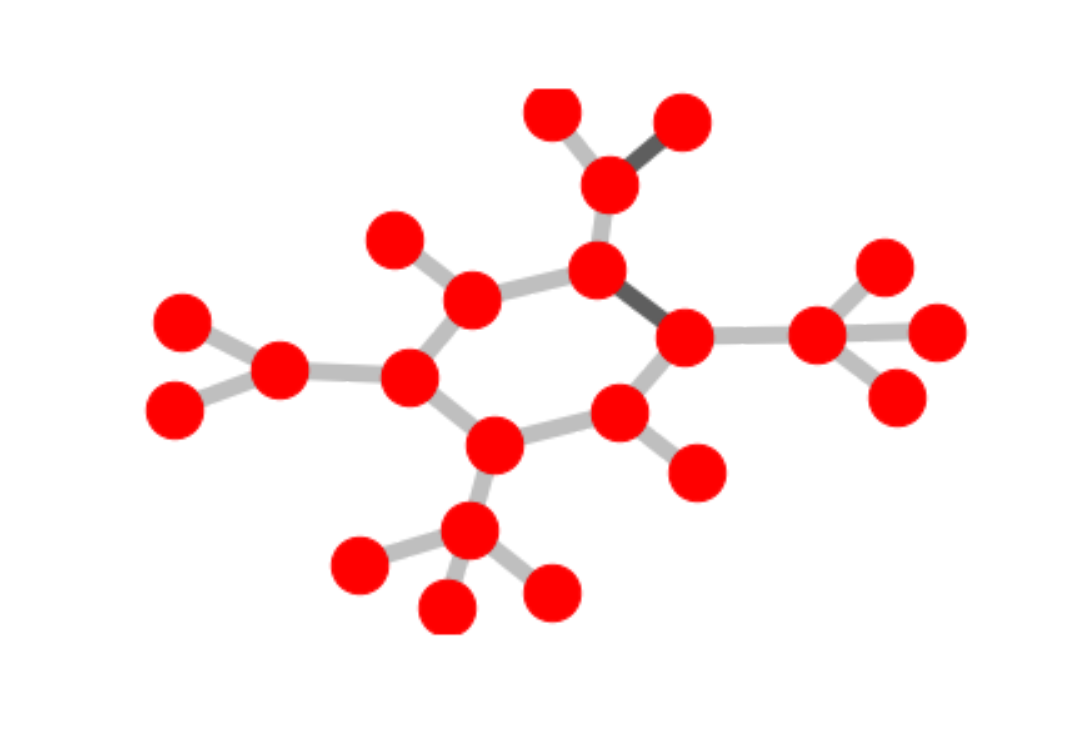} \\

\midrule
\raisebox{3\height}{PGExp} & \includegraphics[scale=0.23]{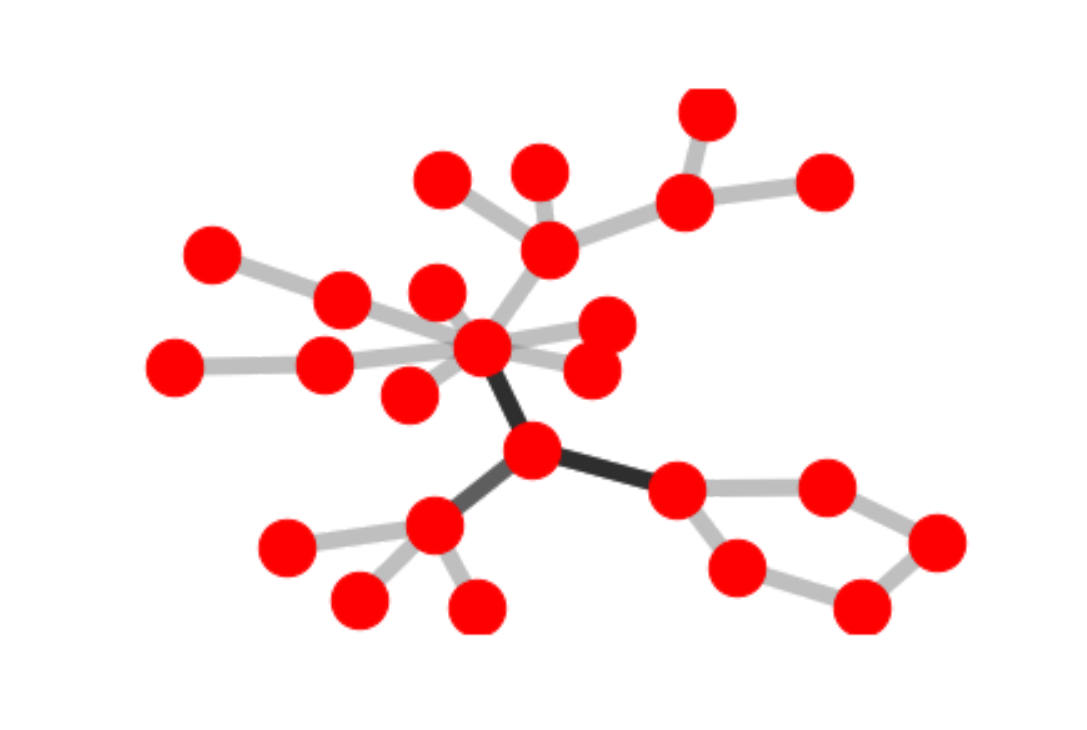} & \includegraphics[scale=0.23]{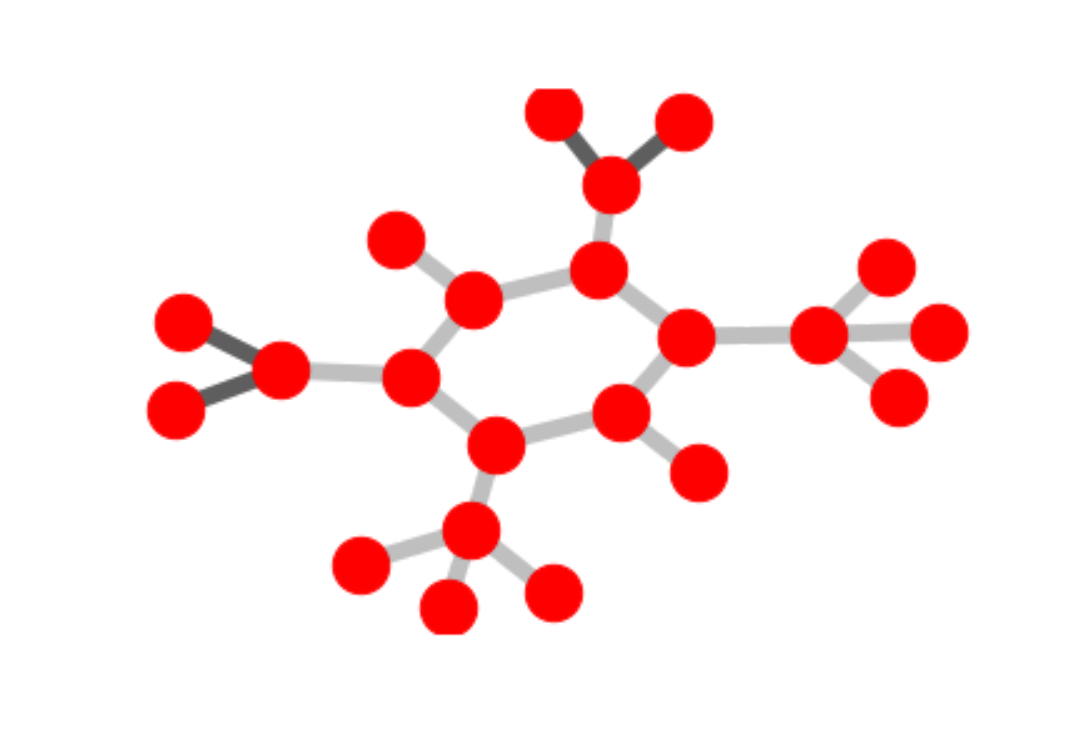} \\
\raisebox{3\height}{MATE+PGExp} & \includegraphics[scale=0.23]{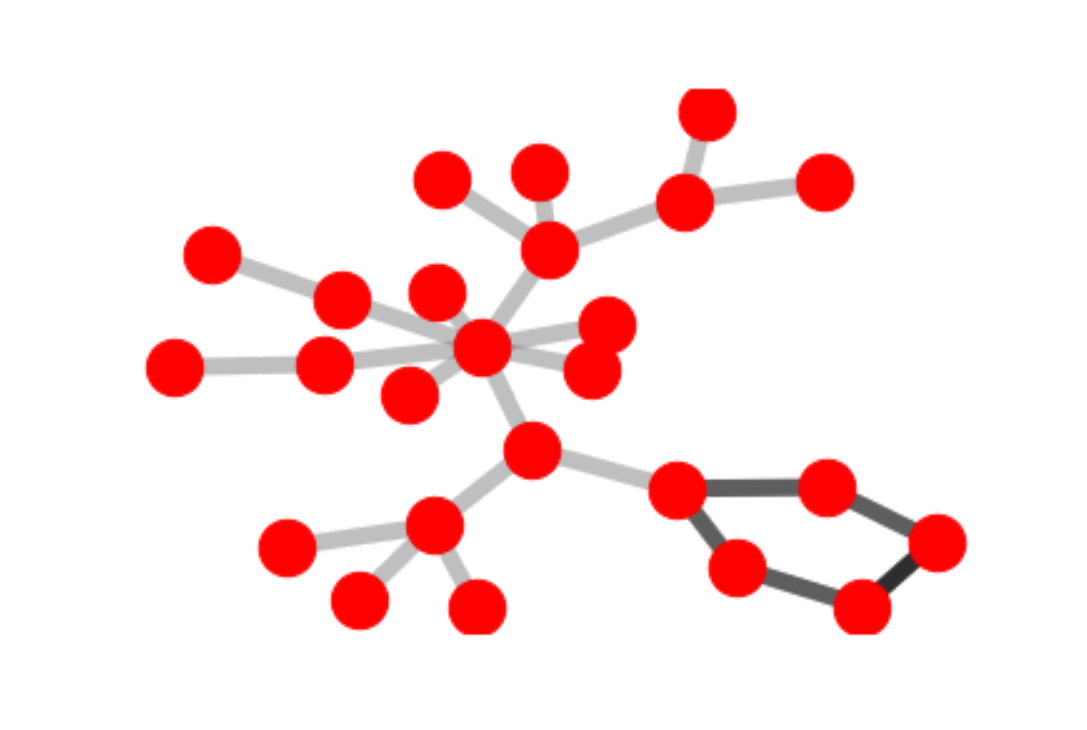} & \includegraphics[scale=0.23]{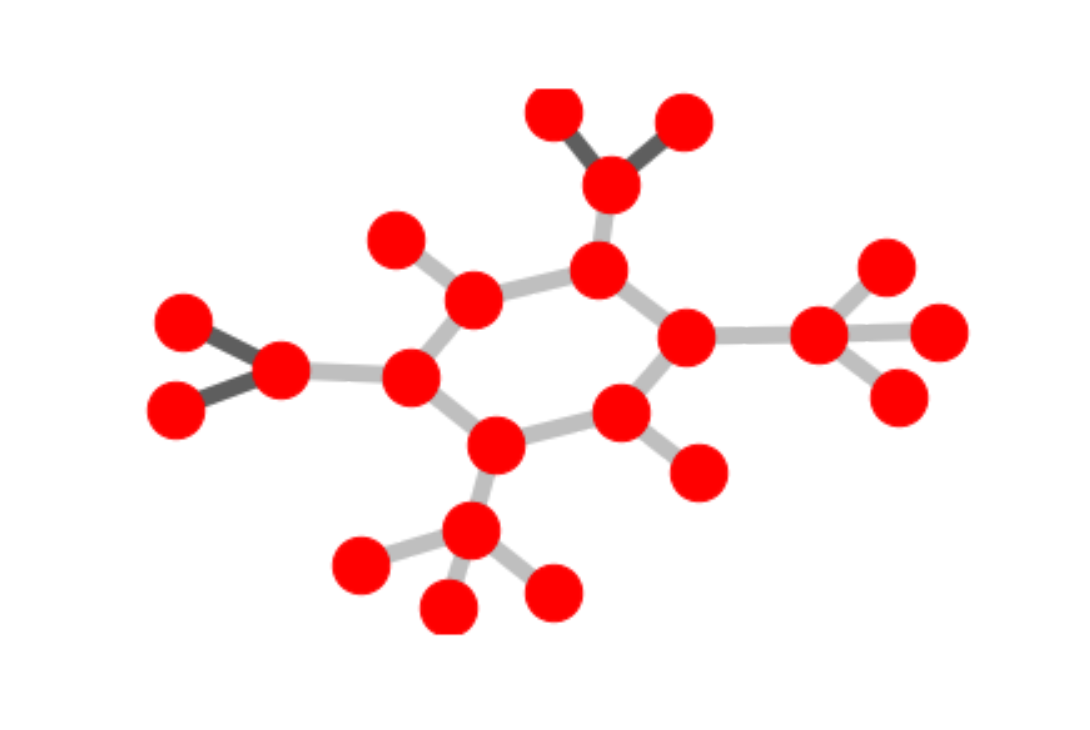} \\
\midrule

\raisebox{3\height}{SubgraphX} & \includegraphics[scale=0.15]{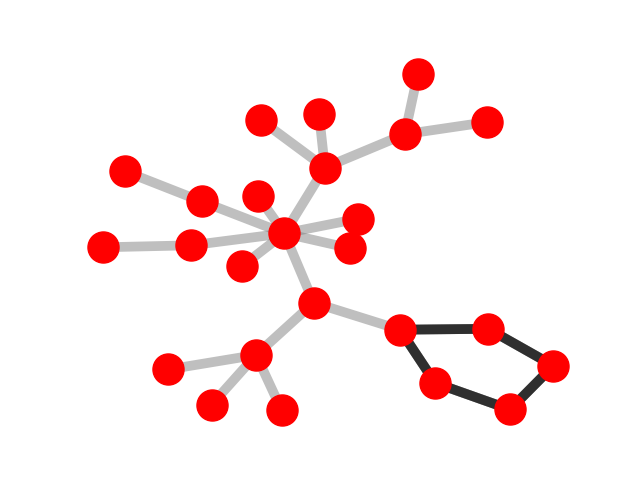} & \includegraphics[scale=0.15]{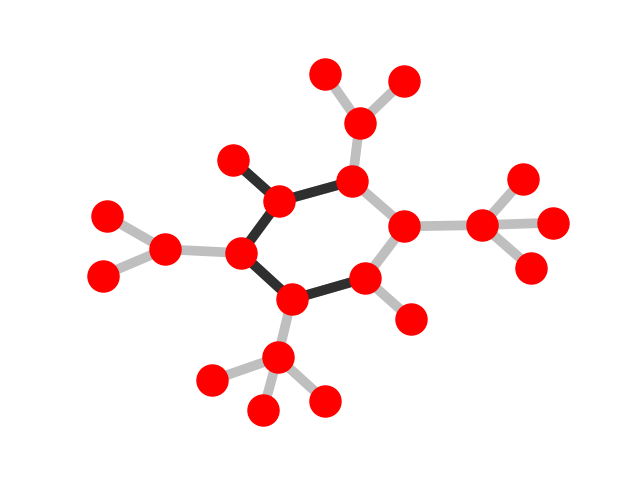} \\
\raisebox{3\height}{MATE+SubgraphX} & \includegraphics[scale=0.15]{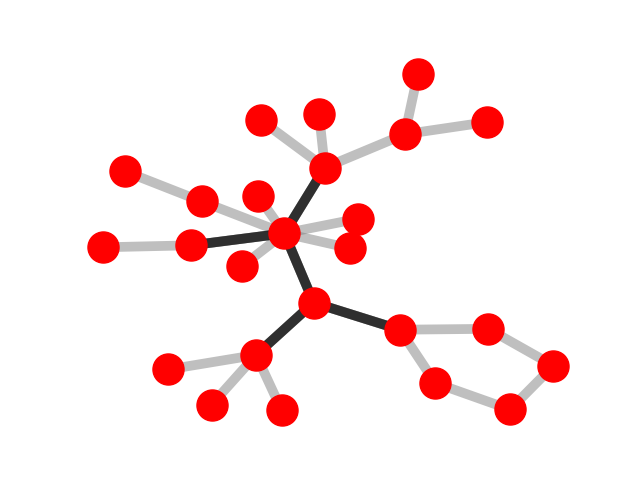} & \includegraphics[scale=0.15]{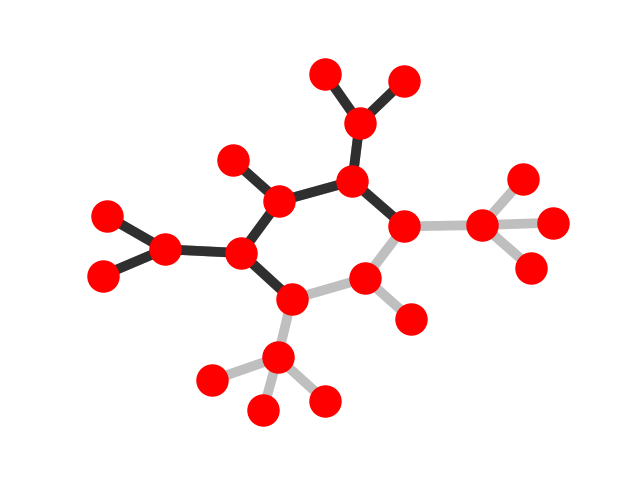} \\
\midrule
\end{tabular}
\label{tab:figPPP}
\end{table}

\subsection{Datasets}
Synthetic datasets are very common in the evaluation of explanation techniques. These datasets contain graph motifs determining the node or graph class. The relationships between the nodes or graphs and their labels are easily understandable by humans. The motifs represent the approximation of the explanation's ground truth. In our evaluation, we consider four synthetic datasets for node classification and 1 for graph classification. BA-shapes generates a base graph with the Barab\'asi-Albert (BA) \cite{albert2002ba} and attaches randomly a house-like, five node motif. It has four labels, one for the base graph, one for the top node of the house motifs, and one for the two upper nodes, followed by the last label for the bottom ones of the house. BA-Community has eight classes and contains 2 BA-shapes graphs with randomly attached edges. The memberships of the BA-shapes graphs and the structural location determine the labels. In Tree-cycle, the base graph is a balanced tree graph with a depth equal to 8. The motifs are a six-node cycle. In this case, we have just two labels, motifs and non-motifs.
Tree-grids substitute the previous motif with a grid of nine nodes. Concerning graph classification, BA-2motifs has 800 graphs and two labels. Each network has a base component generated with the BA model. Then one between the cycle and house-like motif, injected in the graph, determines the resulting label.
All node features are vectors containing all 1s.
The other dataset for graph classification is MUTAG \cite{debnath1991mutag}, a molecular dataset. The dataset contains several molecules represented as graphs where nodes represent atoms and edges chemical bonds. The molecules are labelled based on their mutagenic effect on a specific bacterium. As discussed in \cite{debnath1991mutag} carbon rings with chemical groups NH\textsubscript{2} or NO\textsubscript{2} are present in mutagenic molecules. A good explainer should identify such patterns for the corresponding class. However, the authors of \cite{luo2020parameterized} observed that carbon rings exist in both mutagen and nonmutagenic graphs.

\subsection{Baselines}
We use the same GNN architectures described in \cite{holdijk2021re}. The model has three graph convolutional layers and an additional fully connected classification layer. For node classification, the last layer takes as input the concatenation of the three intermediate outputs. For graph classification, instead, it receives the concatenation of max and mean pooling of the final output.
Concerning the explainers we use GNNExplainer \cite{ying2019gnnexplainer}, PGExplainer \cite{luo2020parameterized} \textcolor{bs}{and SubgraphX \cite{yuan2021x}}. Our baselines will be the explanations provided by the \textcolor{bs}{three} explainers over the outputs of the GNN architecture described previously, trained in a standard fashion. We train the GNN with Adam \cite{kingma2014adam} and an early stopping strategy on a validation split. \textcolor{bs}{GNNExplainer and PGExplainer with the GNNs use the hyperparameters fine-tuned by the authors of \cite{holdijk2021re}. For SubgraphX we used the hyperparameters of the original implementation \cite{yuan2021x}.} We repeat the explanation steps 10 times with different seeds and report in the tables the mean AUC score with the standard deviation.

\subsection{Metrics}
Like in recent works, we divide the evaluation into quantitative and qualitative experiments. For the quantitative part, following \cite{ying2019gnnexplainer,luo2020parameterized,holdijk2021re} we compute the AUC score between the edges inside motifs, considered as positive edges, and the importance weights provided by the explanation methods. Every connection outside the motif has a negative label. High scores for the edges in the ground-truth explanation corresponds to higher explanation accuracy.
The qualitative evaluation, instead, provides a visualisation of the chosen subgraph. Given the mask, we select all the edges that have the weights satisfying a pre-defined threshold. Then, we choose only the nodes that are in a direct subgraph together with the node-to-be-explained. Finally, we select only the top-k edges where k is the number of connections in the motifs. Darker edges have higher weights in the mask than the lighter ones. Nodes are colour coded by their ground-truth label.

\subsection{Hyperparameters}
MATE requires two sets of hyperparameters. The first regards the optimization process with the tuples $(K,\delta)$ and $(T,\alpha)$ and meta step size $\beta$. The tuples drive the explainer training and the adaptation procedure. We set $(K=30,\delta=0.03), \alpha=0.0001$ for all the dataset. Then we set $\beta=0.003$ and $\beta=0.001$ for the node and graph classification dataset respectively. We fine-tuned the number of adaptation steps $T$ for each dataset selecting from the values $[1,3,5,10]$. The second parameter set is the one of GNNExplainer. In this case, we used the same hyperparameters of \cite{holdijk2021re}.

\section{Results}
We investigate the question: Does a model trained to be explainable improves the performances of post-hoc explanation algorithms?

\begin{table*}[th]
\centering
\normalsize
\caption{Explanation accuracy obtained on the model trained with and without our meta-training framework.}
\begin{tabular}{lcccccc}
  & BA-shapes & BA-community & Tree-cycles & Tree-grids & BA-motifs & MUTAG \\
\hline
GNNExp & 0.763 \scriptsize$\pm$ 0.006 & 0.640 \scriptsize$\pm$ 0.004 & 0.479 \scriptsize$\pm$ 0.019 & 0.668 \scriptsize$\pm$ 0.002 & 0.491 \scriptsize$\pm$ 0.004 & 0.637 \scriptsize$\pm$ 0.002 \\
MATE+GNNExp & 0.851 \scriptsize$\pm$ 0.003 & 0.688 \scriptsize$\pm$ 0.004 & 0.523 \scriptsize$\pm$ 0.012 & 0.628 \scriptsize$\pm$ 0.001 & 0.500 \scriptsize$\pm$ 0.001 & 0.680 \scriptsize$\pm$ 0.002 \\
\hline
PGExp & 0.997 \scriptsize$\pm$ 0.001 & 0.868 \scriptsize$\pm$ 0.012 & 0.793 \scriptsize$\pm$ 0.035 & 0.423 \scriptsize$\pm$ 0.012 & 0.101 \scriptsize$\pm$ 0.073 & 0.811 \scriptsize$\pm$ 0.076 \\
MATE+PGExp & \textbf{1.000} \scriptsize$\pm$ 0.000 & \textbf{0.910} \scriptsize$\pm$ 0.008 & \textbf{0.870} \scriptsize$\pm$ 0.011 & \textbf{0.853} \scriptsize$\pm$ 0.010 & \textbf{0.963} \scriptsize$\pm$ 0.020 & \textbf{0.873} \scriptsize$\pm$ 0.100 \\
\hline
SubgraphX & 0.548 \scriptsize$\pm$ 0.002 & 0.473 \scriptsize$\pm$ 0.002 & 0.617 \scriptsize$\pm$ 0.007 & 0.516 \scriptsize$\pm$ 0.006 & 0.610 \scriptsize$\pm$ 0.006 & 0.529 \scriptsize$\pm$ 0.002 \\
MATE+SubgraphX & 0.564 \scriptsize$\pm$ 0.008 & 0.525 \scriptsize$\pm$ 0.001 & 0.642 \scriptsize$\pm$ 0.008 & 0.613 \scriptsize$\pm$ 0.015 & 0.555 \scriptsize$\pm$ 0.011 & 0.576 \scriptsize$\pm$ 0.001 \\
\hline
\end{tabular}
\label{tab:datasets}
\end{table*}

\subsection{Quantitative}
\begin{table*}[th]
\centering
\normalsize
\caption{Model's accuracy with and without our meta-training framework obtained with early stopping. The scores are in the format Train/Validation/Test.}
\begin{tabular}{lcccccc}
  & BA-shapes & BA-community & Tree-cycles & Tree-grids & BA-2motifs & MUTAG \\
    \midrule
Standard & 0.96/1.0/1.0 & 0.85/0.74/0.73 & 0.95/0.98/0.97 & 0.97/0.99/0.98 & 0.99/1.0/0.99 & 0.83/0.84/0.79\\
 \midrule
MATE & 0.97/1.0/1.0 & 0.84/0.70/0.74 & 0.96/0.98/0.97 & 0.97/0.98/0.98 & 0.99/0.99/0.99 & 0.82/0.82/0.78 \\
\hline
	\end{tabular}
	\label{tab:main}
\end{table*}

\textcolor{bs}{In Table \ref{tab:datasets} we report the results in terms of the explanation accuracy obtained using three different explainers on models trained with and without MATE.  We report that in $89\%$ of the cases, MATE-trained models helped the explainers to outscore their counterparts who interpreted standard models. The average increments are $4.6\%$ points on GNNExplainer, $24.6\%$ on PGExplainer and $4.7\%$ on SubgraphX. Part of the success of the combination META+PGExplainer is because the authors of \cite{holdijk2021re} were not able to replicate the original results on Tree-grids. Yet, the results of PGExplainer over the model trained with MATE are comparable with the ones presented in \cite{luo2020parameterized}. We used the same hyper-parameters regardless of the explainers used for the imputation. Therefore, we believe there is still some margin for improvement with a fine-tuning targeting the explainer's accuracy.}

In Table \ref{tab:main} we report the accuracies obtained by the GNN model obtained via standard optimization and with our framework. We show that there are no relevant changes in the utility of the model when optimized to be explainable.

\subsection{Qualitative}

\textcolor{bs}{In this Section, we analyze the qualitative aspect of the explanation subgraphs computed by the post-hoc explainers over a GNN trained with and without our meta-training framework. Since GNNExplainer and PGExplainer output a soft explanation mask, the intensity of the edges in the subgraph reflects the associated confidence. SubgraphX assigns the same importance to each edge being part of the explanation.
We present the results for the node classification task in Table \ref{tab:figPP}. GNNExplainer and PGExplainer provide explanations with darker edges inside the motifs on the MATE-trained models, highlighting greater confidence. Most notably, all the explainers find the cycle motif in Tree-cycles when taking the meta-trained model as input. We report a slight improvement or comparable results for all the other datasets. In Table \ref{tab:figPPP}  we show the interpretations over the graph classification task. In this scenario, PGExplainer is the best performing model among the baselines, but only the variant trained with our meta-training approach is capable of perfectly highlighting both the five node cycle motif and the NH\textsubscript{2} and NO\textsubscript{2} motifs. The combination MATE-SubgraphX on the BA-2motif could improve neither the quantitative nor the qualitative evaluation. However, the same combination correctly includes the ground truth motifs on MUTAG instead of focusing on the carbon ring alone.}

\subsection{Ablation}
We performed an ablation study on the Tree-cycles dataset.
In particular, we observed the change in GNN accuracy and explainability score when varying the number of optimization steps in MATE's inner loop. Table \ref{tab:k} shows the change when acting on the number of GNNExplainer's training steps. We can observe that this value does not influence the GNN accuracy performances. However, we have found a sweet spot in the range $K = [20,50]$ for the explanation scores of both explainers.
Table \ref{tab:t} shows what happens when we perform a different number of adaptation steps $T$ on the `explanation task'. This hyperparameter has a greater impact on both model's accuracy and explainability score. Increasing $T$ worsen the accuracy performance especially for the maximum tested value of $T=10$. Surprisingly, PGExplainer shares this behaviour, meanwhile GNNExplainer performances increase with higher values of $T$. We have found a similar behaviour for all the datasets taken into consideration.

\begin{table}[t]
\centering
\normalsize
\caption{Results of an ablation study on the effect of the number of adaptation steps on the model's accuracy and AUC score. The
ablation study is performed using the Tree-cycles dataset averaging 10 runs.}
\begin{tabular}{l|c|c|c}
K=30 & Tr/Val/Te & PGExplainer  & GNNExplainer \\
 \hline
T=1 & 0.95/0.96/0.95 & \textbf{0.893} \scriptsize$\pm$ 0.009 & 0.495 \scriptsize$\pm$ 0.020 \\
T=3 & 0.95/0.98/0.97 & 0.870 \scriptsize$\pm$ 0.011 & 0.523 \scriptsize$\pm$ 0.012 \\
T=5 & 0.94/0.98/0.94 & 0.832 \scriptsize$\pm$ 0.004 & \textbf{0.554} \scriptsize$\pm$ 0.017 \\
T=10 & 0.90/0.92/0.92 & 0.850 \scriptsize$\pm$ 0.003 & 0.545 \scriptsize$\pm$ 0.018 \\

\end{tabular}
\label{tab:t}
\end{table}

\begin{table}[t]
\centering
\normalsize
\caption{Results of an ablation study on the effect of the number of GNNExplainer optimization steps on the model's accuracy and AUC score. The
ablation study is performed using the Tree-cycles dataset averaging 10 runs.}
\begin{tabular}{l|c|c|c}
T=3 & Tr/Val/Te & PGExplainer & GNNExplainer \\
  \hline
K=10 & 0.95/0.98/0.94 & 0.847 \scriptsize$\pm$ 0.002 & 0.473 \scriptsize$\pm$ 0.016 \\
K=20 & 0.95/0.98/0.97 & 0.850 \scriptsize$\pm$ 0.007 & 0.494 \scriptsize$\pm$ 0.020 \\
K=30 & 0.94/0.98/0.97 & 0.900 \scriptsize$\pm$ 0.006 & 0.542 \scriptsize$\pm$ 0.016 \\
K=40 & 0.96/0.98/0.96 & \textbf{0.918} \scriptsize$\pm$ 0.004 & \textbf{0.547} \scriptsize$\pm$ 0.017 \\
K=50 & 0.96/0.98/0.97 & 0.902 \scriptsize$\pm$ 0.006 & 0.534 \scriptsize$\pm$ 0.016 \\
K=60 & 0.95/0.98/0.94 & 0.838 \scriptsize$\pm$ 0.003 & 0.500 \scriptsize$\pm$ 0.019 \\
K=70 & 0.96/0.98/0.97 & 0.821 \scriptsize$\pm$ 0.004 & 0.546 \scriptsize$\pm$ 0.016 \\
K=80 & 0.95/0.98/0.97 & 0.901 \scriptsize$\pm$ 0.004 & 0.546 \scriptsize$\pm$ 0.016 \\
K=90 & 0.96/0.98/0.97 & 0.838 \scriptsize$\pm$ 0.006 & 0.503 \scriptsize$\pm$ 0.020

\end{tabular}
\label{tab:k}
\end{table}

\section{Conclusions and future works}
\label{sec:conclusions_and_future_works}
In this work, we presented MATE, a meta-learning framework for improving the level of explainability of a GNN at training time. Our approach steers the optimization procedure towards more interpretable minima meanwhile optimizing for the original task. We produce easily processable outputs for downstream algorithms that explain the model's decisions in a human-friendly way. In particular, we optimized the model's parameters to minimize the error of GNNExplainer trained on-the-fly on randomly sampled nodes. Our model-agnostic approach can improve the explanation produced for different GNN architectures by different post-hoc explanation algorithms. Experiments on synthetic and real-world datasets showed that the meta-trained model is consistently easier to explain by GNNExplainer, PGExplainer \textcolor{bs}{and SubgraphX}. A small ablation demonstrated how MATE balances the model's accuracy with the explainability of its outputs. Furthermore, this increase in explainability does not impact the model's prediction performances. Future works may study the feasibility of this approach for other domains like images, audio, and video.

\bibliographystyle{plain}
\bibliography{references}

\begin{IEEEbiography}[{\includegraphics[width=1in,height=1.25in,clip,keepaspectratio]{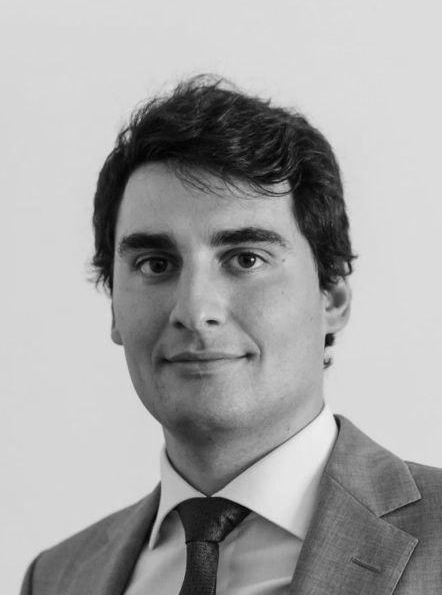}}]{Indro Spinelli}{\space} received a master’s degree in artificial intelligence and robotics in 2019 from Sapienza University of Rome, Italy, where he is currently working towards a PhD in the
Department of Information Engineering, Electronics, and Telecommunications. He is a member of the ``Intelligent Signal Processing and Multimedia'' (ISPAMM) group and his main research interests include graph deep learning, trustworthy machine learning for graph-structured data.
\end{IEEEbiography}

\begin{IEEEbiography}[{\includegraphics[width=1in,height=1.25in,clip,keepaspectratio]{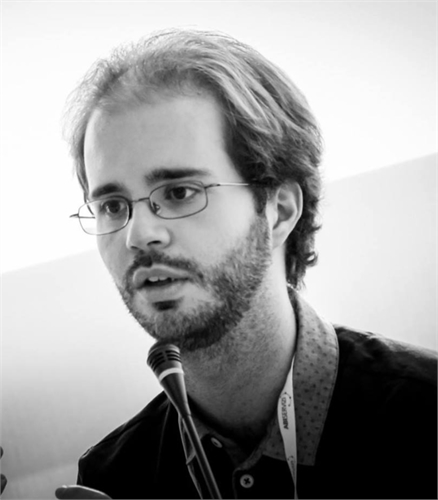}}]{Simone Scardapane}{\space} is currently an Assistant Professor with the Sapienza University of Rome, Italy, where he works on deep learning applied to
audio, video, and graphs, and their application in distributed and decentralized environments. He has authored more than 70 articles in the fields of machine and deep learning. Dr. Scardapane is a member of the IEEE CIS Social Media Sub- Committee, the IEEE Task Force on Reservoir Computing, and the ``Machine Learning in Geodesy" joint-Study group of the International Association of Geodesy. He is Chair of the Statistical Pattern Recognition Techniques TC of the International Association for Pattern Recognition, and Chair of the Italian Association for Machine Learning.
\end{IEEEbiography}

\begin{IEEEbiography}[{\includegraphics[width=1in,height=1.25in,clip,keepaspectratio]{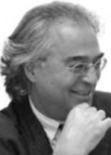}}]{Aurelio Uncini}{\space}(M’88) received the Laurea degree in Electronic Engineering from the University of Ancona, Italy, on 1983 and the Ph.D. degree in Electrical Engineering in 1994 from University of Bologna, Italy. Currently, he is Full Professor with the Department of Information Engineering, Electronics and Telecommunications, where he is teaching Neural Networks, Adaptive Algorithm for Signal Processing and Digital Audio Processing, and where he is the founder and director of the ``Intelligent Signal Processing and Multimedia'' (ISPAMM) group. His present research interests include adaptive filters, adaptive audio and array processing, machine learning for signal processing, blind signal processing and multi-sensors data fusion. He is a member of the Institute of Electrical and Electronics Engineers (IEEE), of the Associazione Elettrotecnica ed Elettronica Italiana (AEI), of the International Neural Networks Society (INNS) and of the Società Italiana Reti Neuroniche (SIREN).
\end{IEEEbiography}
\end{document}